\pdfoutput=1

\documentclass[11pt]{article}

\usepackage[final]{acl}

\usepackage{times}
\usepackage{latexsym}
\usepackage{booktabs}
\usepackage{multirow}
\usepackage{multicol}
\usepackage{amsmath}
\usepackage{makecell}
\usepackage{enumitem}
\usepackage{xcolor}
\usepackage{float}
\usepackage{cleveref}

\usepackage[T1]{fontenc}

\usepackage[utf8]{inputenc}

\usepackage{microtype}

\usepackage{inconsolata}

\usepackage{graphicx}

\crefname{table}{Table}{Tables}

\title{Large Language Models for Multilingual Previously Fact-Checked\\Claim Detection}

\author{Ivan Vykopal$^{1, 2}$, Matúš Pikuliak$^{2}$, Simon Ostermann$^{3, 4}$, Tatiana Anikina$^{3}$,\\\textbf{Michal Gregor}$^{2}$ \and \textbf{Marián Šimko}$^{2}$ \\
    $^{1}$ Faculty of Information Technology, Brno University of Technology, Brno, Czech Republic\\
    $^{2}$ Kempelen Institute of Intelligent Technologies, Bratislava, Slovakia \\
    \texttt{\{name.surname\}@kinit.sk} \\
    $^{3}$ German Research Center for Artificial Intelligence (DFKI), Saarbrücken, Germany  \\
    \texttt{\{name.surname\}@dfki.de}\\
    $^{4}$ Centre for European Research in Trusted AI (CERTAIN)
  }

\begin{document}
\maketitle
\begin{abstract}

In our era of widespread false information, human fact-checkers often face the challenge of duplicating efforts when verifying claims that may have already been addressed in other countries or languages. As false information transcends linguistic boundaries, the ability to automatically detect previously fact-checked claims across languages has become an increasingly important task. This paper presents the first comprehensive evaluation of large language models (LLMs) for \textit{multilingual} previously fact-checked claim detection. We assess seven LLMs across 20 languages in both monolingual and cross-lingual settings. Our results show that while LLMs perform well for high-resource languages, they struggle with low-resource languages. Moreover, translating original texts into English proved to be beneficial for low-resource languages. These findings highlight the potential of LLMs for multilingual previously fact-checked claim detection and provide a foundation for further research on this promising application of LLMs.
\end{abstract}

\section{Introduction}

False information spreads rapidly across digital platforms, undermining public trust and distorting public discourse, posing a significant challenge to society. In response, \textit{fact-checking} has become a crucial defense against false information. However, manual fact-checking remains a slow and labor-intensive process, often lagging behind the speed at which false claims propagate. To scale these efforts, automated fact-checking has emerged as a promising avenue to assist fact-checkers~\cite{doi:10.1126/science.aap9559, aimeur2023fake}.

\begin{figure}
\centering
\hspace*{-0.25cm}
\includegraphics[width=1.07\columnwidth]{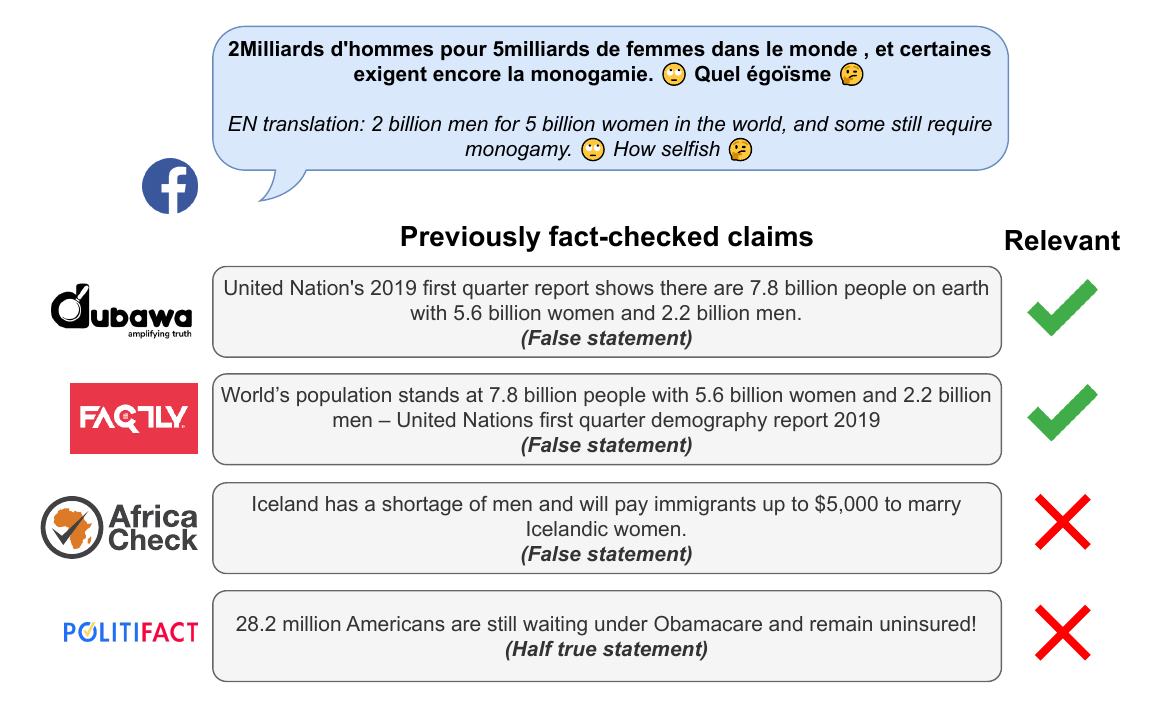}
\caption{An example of a Facebook post with four previously fact-checked claims retrieved by the multilingual E5 embedding model, annotated by human annotators for relevance. Two claims are relevant to the post, while two are irrelevant.}
\label{fig:example}
\end{figure}

A practically important but underexplored area of automated fact-checking is \textit{previously fact-checked claim detection} (PFCD), the task of identifying whether a given claim has already been verified (see an example in Figure~\ref{fig:example}). Misinformation often reappears in paraphrased or translated forms in different languages, leading to repeated verification of the same claims~\cite{ barroncedeno2020overviewcheckthat2020automatic, nakov2021automatedfactcheckingassistinghuman, 10.1145/3512974}. Interviews with professional fact-checkers and journalists confirm that re-verifying already fact-checked claims is inefficient, which makes previously fact-checked claim detection one of the most crucial tasks~\cite{hrckova2024autonomationautomationactivitiesneeds}.

While PFCD is crucial in combating misinformation, its challenges become even more pronounced in multilingual contexts~\cite{vykopal2024generativelargelanguagemodels}. With hundreds of languages and independent fact-checking organizations worldwide, fact-checkers increasingly need to identify previously fact-checked claims across languages to avoid duplicating work. Yet, most existing PFCD approaches are monolingual, leaving fact-checkers in low-resource languages at a disadvantage when combating false information. Moreover, determining whether two claims refer to the same fact becomes increasingly challenging when they are in different languages and cultural contexts, especially in cases where high-quality data are limited. These challenges underscore the need for multilingual approaches that can operate reliably across languages and bridge the gap between multilingual misinformation and monolingual fact-checking resources.


In this work, we build upon the definitions of \citet{pikuliak-etal-2023-multilingual} and frame PFCD as a binary classification task to better reflect the practical needs of fact-checkers. We investigate whether large language models can reason about semantic relationships between claims and posts, aiming to improve the detection of previously fact-checked claims in multilingual and low-resource scenarios.

To this end, we conducted \textbf{the first systematic evaluation of multilingual LLMs for PFCD}, treating the task as binary relevance classification. We benchmark seven open-source LLMs across 20 languages, covering both monolingual and cross-lingual settings.\footnote{Code and data are available at: \url{https://github.com/kinit-sk/llms-claim-matching}} To evaluate the capabilities of multilingual LLMs on the PFCD task, we created a novel, manually annotated multilingual dataset consisting of 16K post-claim pairs labeled for relevance. In addition, we provide empirical insights into the various prompting strategies to understand how task framing affects LLM performance and analyze where current models lag behind, especially in low-resource languages. Our findings highlight the potential and limitations of LLMs, offering a foundation for future research.


Our contributions are twofold:

\begin{itemize}
\item We conduct the \textbf{first comprehensive evaluation of LLMs for PFCD}, analyzing their performance in monolingual and cross-lingual scenarios and the effectiveness of various prompting techniques.
\item We introduce a \textbf{novel manually annotated multilingual dataset}, comprising 16K post-claim pairs that assess the relevance between social media posts and fact-checked claims, supporting the research in multilingual fact-checking.
\end{itemize}

\section{Related Work}

\paragraph{Previously Fact-Checked Claim Detection.} 

Detecting previously fact-checked claims, also known as claim-matching~\cite{kazemi2021claimmatchingenglishscale}, aims to identify relevant claims for a given input~\cite{shaar2020knownliedetectingpreviously}. This approach helps to reduce the need to revisit previously fact-checked information. Traditionally, the PFCD is framed as a retrieval task, where systems rank previously fact-checked claims based on their similarity to an input claim, using embedding-based or IR-based methods~\cite{kazemi2022matchingtweetsapplicablefactchecks, larraz2023semantic}. Most studies have focused on monolingual settings, primarily in English~\cite{shaar2020knownliedetectingpreviously, shaar2022assistinghumanfactcheckersdetecting, hardalov2022crowdcheckeddetectingpreviouslyfactchecked}, limiting their applicability to other languages.

Recent efforts have begun exploring multilingual PFCD by developing multilingual datasets and exploring the performance of existing IR systems~\cite{kazemi2021claimmatchingenglishscale}. \citet{pikuliak-etal-2023-multilingual} created the \textbf{\textit{MultiClaim}} dataset with over 27 languages, demonstrating that English embedding models with translated data achieved superior results compared to multilingual models with input in the original language.

\paragraph{LLMs for Detecting Previously Fact-Checked Claims.}

The rise of LLMs has opened new possibilities for PFCD in both monolingual and multilingual settings. While most existing research relies on embedding-based similarity, only a few studies have applied LLMs to PFCD, and these are limited to single-language scenarios~\cite{vykopal2024generativelargelanguagemodels}. Two main strategies dominate prior LLM-based approaches: \textit{textual entailment}~\cite{10.1145/3589335.3651910, 10.1145/3589335.3651504}, which labels claim relationships as entailment, contradiction or neutral, and \textit{generative re-ranking}~\cite{shliselberg2022riet, neumann-etal-2023-deep}, which re-ranks retrieved fact-checks based on conditional probabilities.

These approaches do not fully align with the needs of fact-checkers. Entailment labels or ranked outputs do not always capture the broader notion of \textit{relevance}. Our work builds on this gap by proposing a broader framing of PFCD. We formulate PFCD as a binary classification problem, where the goal is to determine whether a fact-checked claim is relevant to a given post.

Our definition of \textit{relevance} encompasses a wider range of relationships, including entailment, contradiction, partial overlap and generalization. Unlike previous work, we define \textit{relevance} less strictly without considering the stance of the post or retrieved claim: a fact-checked claim is considered relevant if it can assist in verifying the post, regardless of whether it supports or refutes the claim. This definition aligns more closely with human fact-checking workflows and emphasizes shared information useful for verification.

For example, consider the post: \textit{"This is a recent photo of the peasant protests in the Netherlands. German media do not report on the current protests."} and the previously fact-checked claim: \textit{"This is what they don't show you Port of Rotterdam, Thursday 18 November flat, strikes Media has plenty of time for riots"}. 

While there is no direct entailment or contradiction between them, the fact-checked claim contains information relevant to verifying the post, illustrating our broader notion of relevance. See detailed explanation in Appendix~\ref{sec:relevance}.



\section{Methodology}
\label{sec:methodology}

\begin{figure}
\centering
\hspace*{-0.25cm}\includegraphics[width=1.07\columnwidth]{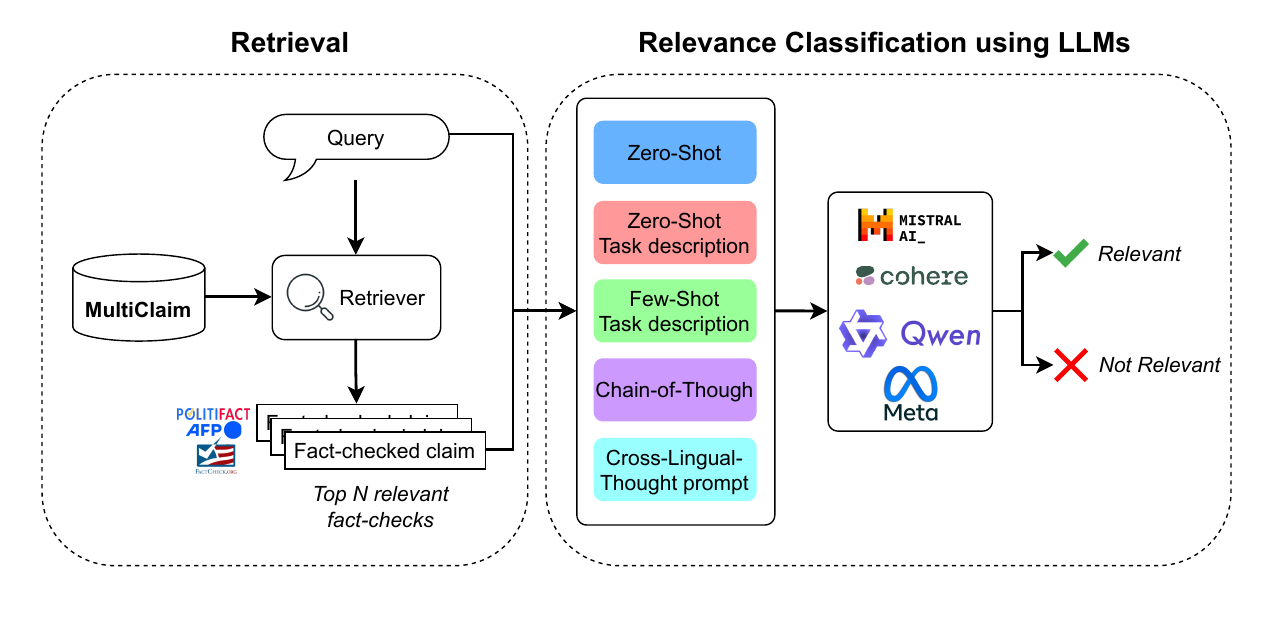}
\caption{Our PFCD pipeline, consisting of (1) a retrieval of the top N most similar previously fact-checked claims (left-hand side) and (2) a classification of the relevance between social media posts and fact-checked claims using LLMs (right-hand side).}
\label{fig:pipeline}
\end{figure}

We assess the ability of LLMs to determine the relevance between social media posts and previously fact-checked claims by instructing them to classify each post-claim pair as either relevant or irrelevant. Our experiments consider both \textit{monolingual settings}, where the post and fact-checked claim are in the same language, and \textit{cross-lingual settings}, where they are in different languages. 

We proposed a pipeline, illustrated in Figure~\ref{fig:pipeline}, to facilitate the evaluation by identifying fact-checked claims relevant to a given social media post. The pipeline consists of two main steps. First, the retriever component retrieves the \textit{N} most similar previously fact-checked claims from a database using an embedding-based similarity (Section~\ref{sec:dataset}). In the second step, an LLM determines the relevance of the retrieved claims to the social media post. The role of LLM is, therefore, to filter out false positives from the first stage. 
To validate the pipeline, we created a manually annotated dataset, where human annotators assessed the relevance between posts and retrieved claims (Section~\ref{sec:annotation}). 

\subsection{Dataset}
\label{sec:dataset}

We evaluate LLM capabilities on the PFCD task using the \textit{\textbf{MultiClaim}} dataset~\cite{pikuliak-etal-2023-multilingual}, which comprises 206K fact-checks in 39 languages and 28K social media posts in 27 languages, with 31K pairings between fact-checks and posts. Pairs of social media posts and fact-checks were collected based on annotations made by professional fact-checkers, who reviewed the posts and linked them to appropriate fact-checks. These data were sourced directly from fact-checks, which specify the social media posts they address. Since each fact-check typically covers only a few posts related to the target claim, there are many potentially correct pairings between posts and fact-checks that are not annotated. In other words, the annotations are not \textit{exhaustive}, making it impossible to measure recall and allowing only a precision-based evaluation. In our experiments, we considered 20 languages with at least 100 posts each, selecting representative subsets for each language.

\subsection{Human Annotation}
\label{sec:annotation}

To ensure a comprehensive evaluation, we aim to address the lack of exhaustiveness in \textbf{\textit{MultiClaim}} as much as possible. Approximating retrieval accuracy requires a complete annotation, but measuring recall directly is infeasible, as it would require comparing each post to every claim in the database. Instead, we approximate recall by retrieving and annotating a representative subset of the data. While this selection is biased toward the retriever and does not allow for exact recall measurement, we believe it provides the most fair evaluation. The quality of the retrieval part is detailed in Appendix~\ref{app:retrieval-quality}.



\paragraph{Data Selection.}

We selected 20 languages from diverse families and scripts to ensure broad linguistic coverage. For monolingual settings, we selected 40 posts per language and retrieved their top 10 fact-checked claims in the same language using the \textit{Multilingual E5 Large} model~\cite{wang2024multilinguale5textembeddings}.

For cross-lingual settings, we defined 20 language pairs, incorporating a variety of language combinations (e.g., Slovak posts with English fact-checks). For each post, we retrieved the top 100 fact-checked claims in languages different from the post's language. From these, we selected 400 post-claim pairs per language combination.

Our dataset, \textbf{\textit{AMC-16K}} (\textit{Annotated MultiClaim-16K}), consists of 8K monolingual and 8K cross-lingual pairs, as detailed in Table~\ref{tab:analyzed-languages} in Appendix~\ref{app:dataset}.


\begin{figure*}[t]
\centering
\includegraphics[width=0.8\linewidth]{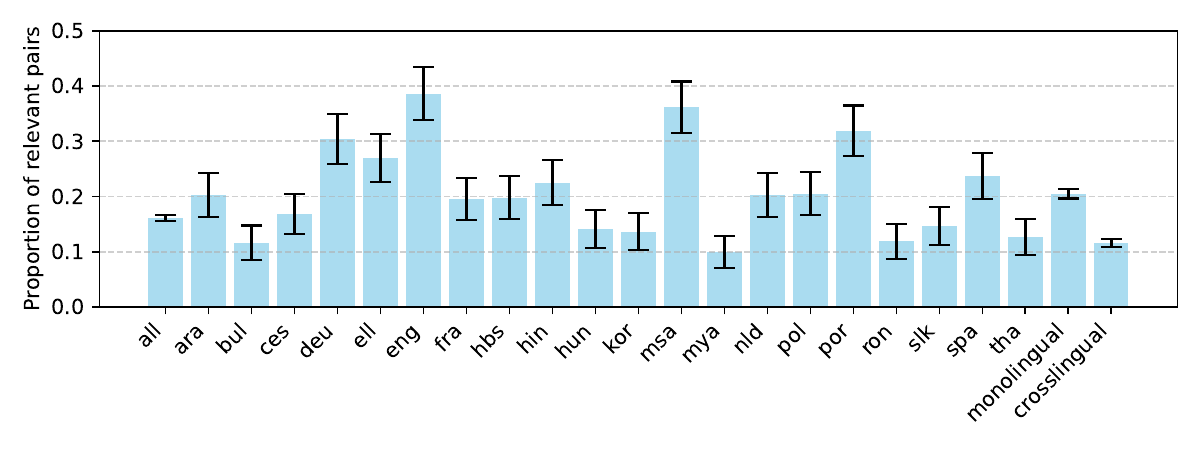}
\caption{The proportion of relevant pairs among 400 annotated samples per language in monolingual and cross-lingual settings. Confidence intervals were computed using the Agresti-Coull method~\cite{Agresti01051998}.}
\label{fig:data-stats}
\end{figure*}

\paragraph{Annotation.}

Six annotators evaluated the relevance of 16K claim-post pairs. For each pair, they assessed the relevance between the post and fact-checked claim as relevant (\textit{Yes}), irrelevant (\textit{No}) or \textit{Cannot tell} according to guidelines we publish alongside this paper. Following the initial annotation, all cases marked as \textit{cannot tell} were reviewed and re-categorized into \textit{Yes} and \textit{No} categories by one of the authors. While each pair received a single annotation due to the dataset size, we implemented two agreement evaluations. First, a pre-annotation alignment test with all annotators to assess their understanding of the guidelines, yielding a \textit{Fleiss' kappa} score of 0.60 (moderate agreement). Second, the four most active annotators completed a post-annotation test, which resulted in a score of 0.62 (substantial agreement), confirming sufficient consistency of our methodology. More details on human annotation can be found in Appendix~\ref{app:human-annotation}.

Annotation results for languages and settings (monolingual vs. cross-lingual) are shown in Figure~\ref{fig:data-stats}. Overall, 16\% of pairs were labeled relevant, with the rest classified as irrelevant. In languages like \textit{English}, \textit{Malay}, \textit{Portuguese}, and \textit{German}, the proportion of relevant pairs exceeded 30\%.



\subsection{Experimental Setup}
\label{sec:experimental-setup}

To assess LLMs' ability to identify the relevance between posts and fact-checked claims, we leveraged our \textit{\textbf{AMC-16K}} dataset, four baselines, seven LLMs and five prompting strategies.

\paragraph{Baselines.}

As baselines, we use two text embedding models: the \texttt{Multilingual E5 Large} and the English-only \texttt{GTR-T5 Large}. We convert semantic similarity scores between posts and fact-checked claims to binary labels using thresholds optimized for Youden's Index~\cite{youden-index}. 

In some sense, our task is similar to Natural Language Inference (NLI): if a post is entailed or paraphrased by a fact-checked claim, it can be considered relevant. Given this connection, we included two NLI models as baselines, \texttt{DeBERTa v3 Large}\footnote{\url{https://huggingface.co/MoritzLaurer/DeBERTa-v3-large-mnli-fever-anli-ling-wanli}} and \texttt{mDeBERTa v3 Base}\footnote{\url{https://huggingface.co/MoritzLaurer/mDeBERTa-v3-base-mnli-xnli}}. We classify NLI relations between posts and fact-checked claims, treating \textit{entailment} relations as relevant and all other labels as irrelevant (see Appendix~\ref{app:entailment}).

\paragraph{Large Language Models.}

Based on preliminary experiments (see Appendix~\ref{app:preliminary}), we selected the top three open-source LLMs with less than 10B parameters, referred to hereafter as \textit{10B- LLMs}, and four LLMs with more than 70B parameters, referred to as \textit{70B+ LLMs}. To optimize resource efficiency for 70B+ LLMs, we employed their quantized versions. Table~\ref{tab:models} lists all LLMs used in our experiments.

\begin{table}
\resizebox{\columnwidth}{!}{%
\begin{tabular}{llrll}
\toprule
\textbf{Model} & \textbf{\# Params} & \multicolumn{1}{l}{\textbf{\# Langs}} & \textbf{Organization} & \textbf{Citation} \\
\midrule
\texttt{Mistral Large} & 123 B & 11 & Mistral AI & \citet{mistral-large} \\ 
\texttt{C4AI Command R+} & 104 B & 23 & Cohere For AI & \citet{cohere_for_ai_2024} \\
\texttt{Qwen2.5 Instruct} & 72 B & 29 & Alibaba & \citet{yang2024qwen2technicalreport} \\
\texttt{Llama3.1 Instruct} & 70 B & 8 & Meta & \citet{grattafiori2024llama3herdmodels} \\
\midrule
\texttt{Llama3.1 Instruct} & 8 B & 8 & Meta & \citet{grattafiori2024llama3herdmodels} \\
\texttt{Qwen2.5 Instruct} & 7 B & 29 & Alibaba & \citet{yang2024qwen2technicalreport} \\
\texttt{Mistral v3} & 7 B & 1 & Mistral AI & \citet{jiang2023mistral7b} \\
\bottomrule
\end{tabular}}
\caption{A list of models evaluated on the task of detecting previously fact-checked claims.}
\label{tab:models}
\end{table}

\paragraph{Prompting Strategies.}

We investigated five strategies for instructing LLMs to identify relevant claim-post pairs: (1) \textit{zero-shot}; (2) \textit{zero-shot with task description}; (3) \textit{few-shot with task description}; (4) \textit{chain-of-thought}; and (5) \textit{cross-lingual-thought prompting}. These strategies were shown to be effective in prior research~\cite{NEURIPS2020_1457c0d6, huang-etal-2023-languages}. Rather than conducting extensive prompt engineering, we aim to assess LLM performance under realistic, minimally tuned conditions. All prompt templates used are aligned with our human annotation guidelines. Examples of our prompt templates are shown in Figure~\ref{fig:prompts}. The prompt formulations were chosen through preliminary experiments evaluating multiple prompts on the annotated data from~\citet{pikuliak-etal-2023-multilingual}.

In \textbf{\textit{Zero-shot prompting}} setup, the LLM is presented with a claim-post pair and the question \textit{"Is the claim relevant to the social media post?"} without additional context or instruction. This serves as a baseline for evaluating the LLM's ability to infer relevance based solely on the input texts. In contrast, the \textbf{\textit{zero-shot with task description}} approach enhances original zero-shot settings by providing a task description in the system prompt, derived from our human annotation guidelines, to clarify the relevance task. 

For the \textbf{\textit{few-shot with task description}} strategy, we include the same task description along with ten task-specific demonstrations (five relevant, five irrelevant, using random order) selected from a subset of manually annotated data from~\citet{pikuliak-etal-2023-multilingual}. These demonstrations were selected based on cosine similarity between social media posts and fact-checked claims. More details on the selection process of demonstrations are in Appendix~\ref{app:fewshot}.

Recognizing the importance of reasoning in fact-checking, we adopt \textbf{\textit{chain-of-thought}} (CoT) prompting~\cite{10.5555/3600270.3602070}, which guides LLMs to reason through the relevance decision step-by-step using the phrase \textit{"Let's think step by step"}. By encouraging intermediate reasoning, CoT aims to improve the understanding of posts and their relationship to fact-checked claims. 

Finally, \textbf{\textit{cross-lingual-thought prompting}} (XLT)~\cite{huang-etal-2023-languages}, addresses multilingual input by guiding the model to translate the input into English before performing the downstream task. In our case, we instruct the model to translate both claims and posts into English and then determine the relevance between them, while also using the task description. This strategy leverages the generally stronger performance of LLMs in English and reduces potential cross-lingual noise.


\subsection{Evaluation}

We evaluate the capabilities of LLMs for PFCD as a binary classification, aiming to determine the relevance of fact-checked claims and a given post. For evaluation, we leverage \textit{Macro F1} due to the fact that the annotated dataset is inherently imbalanced. In addition, we calculate \textit{True Negative Rate} (TNR), reflecting the proportion of irrelevant pairs correctly filtered, and the \textit{False Negative Rate} (FNR), indicating how many relevant pairs were incorrectly identified as irrelevant.

\section{Experiments and Results}

This section presents overall findings on LLMs' performance for the PFCD task (Section~\ref{sec:overall_results}), followed by evaluations in \textit{monolingual} (Section~\ref{sec:mono}) and \textit{cross-lingual} settings (Section~\ref{sec:cross}). We also assess the impact of English translations -- provided in the \textit{\textbf{MultiClaim}} dataset via the Google Translate API -- on LLMs' performance (Section~\ref{sec:english}). Since the translations are provided with the dataset, we do not compare different translation models or evaluate translation quality. For calculating statistical significance, we used the Mann-Whitney U test~\cite{497e1044-d5b0-30a9-b230-3ca0f10d6f6c} for pairwise comparisons and the Kruskal-Wallis test~\cite{Kruskal01121952} for multiple groups, with significance defined as $p < 0.05$.

\begin{figure*}[t]
\centering
\includegraphics[width=0.85\textwidth]{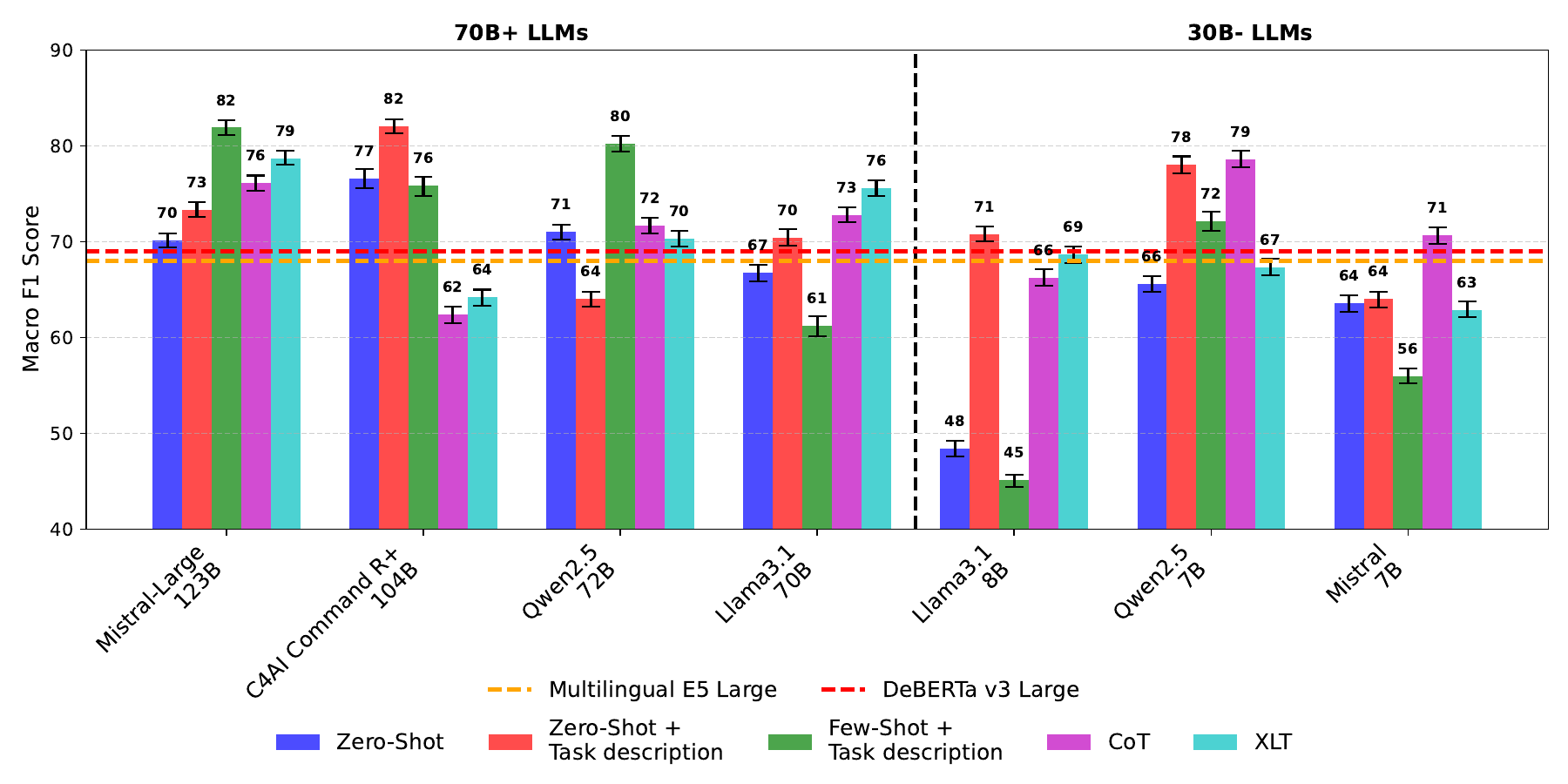}
\caption{Performance comparison of LLMs across five prompting strategies in the original language, measured by Macro F1 score with confidence intervals. Horizontal lines indicate the best-performing baselines. 
}
\label{fig:overall-performance}
\end{figure*}

\begin{figure*}[t]
\centering
\includegraphics[width=\textwidth]{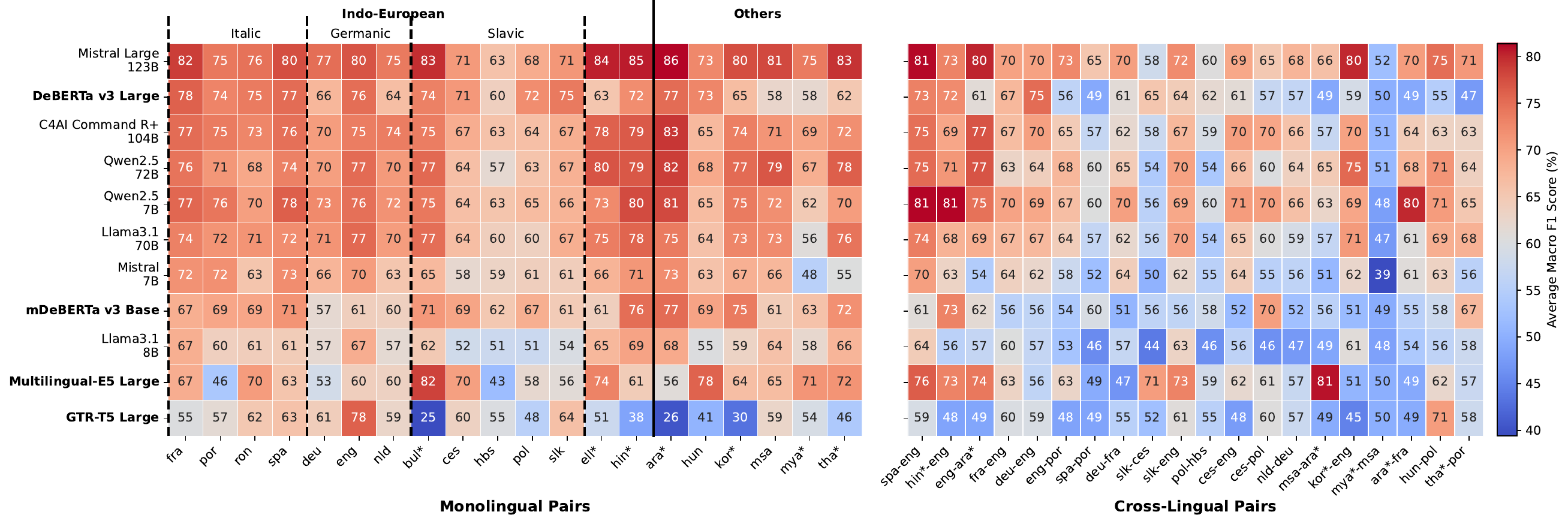}
\caption{Performance of 70B+ and 10B- LLMs across 20 individual languages (left-side) and 20 cross-lingual combinations (right-side). The average Macro F1 performance for each LLM is calculated across all prompting strategies. Languages marked with * use a non-Latin script. \texttt{Mistral Large} demonstrates strong performance across both individual languages and cross-lingual combinations. Baseline models are indicated in \textbf{bold}.}
\label{fig:large_models_languages}
\end{figure*}

\subsection{Overall Assessment}
\label{sec:overall_results}

Figure~\ref{fig:overall-performance} presents the overall results. LLMs generally show strong performance in identifying relevant fact-checked claims across languages, with top LLMs achieving Macro F1 above 80\%. However, performance varies notably by model size, prompting strategy, and language. \textbf{70B+ LLMs consistently outperformed their smaller counterparts} (statistically significant; $p < 0.05$), with \texttt{Mistral Large} and \texttt{C4AI Command R+} emerging as particularly effective. Many LLMs and prompting strategies surpassed the baselines, while \texttt{Llama3.1 8B} and \texttt{Mistral 7B} lagged behind in most strategies.

The effectiveness of prompting strategies depends on the LLM's size and training. \textbf{For 70B+ LLMs, few-shot prompting yields the best results} for \texttt{Mistral Large} and \texttt{Qwen2.5 72B} ($p<0.05$), suggesting these LLMs can effectively leverage demonstrations to understand the task and enable them to leverage context effectively. In contrast, \textbf{10B- LLMs perform better with CoT prompting}, indicating they benefit from the reasoning (e.g., \texttt{Qwen2.5 7B}); this was statistically significant ($p<0.05$) compared to other techniques, except Zero-Shot + Task description, where no significant difference was found. CoT also demonstrated strong performance for LLMs with advanced capabilities, such as \texttt{Llama3.1} and \texttt{Mistral Large}.

\begin{table*}[t]
\tiny
\centering
\begin{tabular*}{\textwidth}{@{\extracolsep{\fill}}lc|cc|cc|cc|cc|cc|cc@{}}
\toprule
\multicolumn{1}{c}{\multirow{2}{*}{\textbf{Model}}} & \multicolumn{1}{c|}{\multirow{2}{*}{\textbf{Version}}} & \multicolumn{2}{c|}{\textbf{Zero-Shot}} & \multicolumn{2}{c|}{\textbf{\makecell[c]{Zero-Shot +\\Task Description}}} & \multicolumn{2}{c|}{\textbf{\makecell[c]{Few-Shot +\\Task Description}}} & \multicolumn{2}{c|}{\textbf{CoT}} & \multicolumn{2}{c|}{\textbf{XLT}} & \multicolumn{2}{c}{\textbf{Average}} \\
\cline{3-14}
& & \textit{Mono} & \textit{Cross} & \textit{Mono} & \textit{Cross} & \textit{Mono} & \textit{Cross} & \textit{Mono} & \textit{Cross} & \textit{Mono} & \textit{Cross} & \textit{Mono} & \textit{Cross} \\
\midrule
\multicolumn{10}{l}{\quad \textit{Baselines}} \\
\midrule
\multirow[t]{2}{*}{\texttt{Multilingual E5 Large}} & Og & 64.92 & \textbf{65.50} & 64.92 & \textbf{65.50} & 64.92 & \textbf{65.50} & 64.92 & \textbf{65.50} & 64.92 & \textbf{65.50} & 64.92 & \textbf{65.50} \\
& En & \underline{77.53} & \underline{68.45} & \underline{77.53} & \underline{68.45} & \underline{77.53} & \underline{68.45} & \underline{77.53} & \underline{68.45} & \underline{77.53} & \underline{68.45} & \underline{77.53} & \underline{68.45} \\
\texttt{GTR-T5 Large} & En & 75.52 & 67.88 & 75.52 & 67.88 & 75.52 & 67.88 & 75.52 & 67.88 & 75.52 & 67.88 & 75.52 & 67.88 \\
\midrule
\texttt{DeBERTa v3 Large (NLI)} & Og & \textbf{70.32} & 64.51 & \textbf{70.32} & 64.51 & \textbf{70.32} & 64.51 & \textbf{70.32} & 64.51 & \textbf{70.32} & 64.51 & \textbf{70.32} & 64.51 \\
 & En & 74.05 & 68.11 & 74.05 & 68.11 & 74.05 & 68.11 & 74.05 & 68.11 & 74.05 & 68.11 & 74.05 & 68.11 \\
\texttt{mDeBERTa v3 Base (NLI)} & Og & 67.38 & 58.68 & 67.38 & 58.68 & 67.38 & 58.68 & 67.38 & 58.68 & 67.38 & 58.68 & 67.38 & 58.68 \\
 & En & 64.84 & 59.05 & 64.84 & 59.05 & 64.84 & 59.05 & 64.84 & 59.05 & 64.84 & 59.05 & 64.84 & 59.05 \\
\midrule
\multicolumn{14}{l}{\quad \textit{LLMs with more than 70B parameters (70B+ LLMs)}} \\
\midrule 
\multirow[t]{2}{*}{\texttt{Mistral Large 123B}} & Og & 72.06 & 67.09 & 74.20 & 71.33 & \textbf{82.46} & \textbf{80.54} & \textbf{79.40} & \textbf{71.60} & \textbf{81.98} & \textbf{74.29} & \textbf{78.02} & \textbf{72.97} \\
 & En & 75.50 & 67.41 & 79.39 & 72.38 & 81.64 & \underline{78.06} & \underline{78.95} & \underline{72.09} & - & - & \underline{78.87} & \underline{72.49} \\
\multirow[t]{2}{*}{\texttt{C4AI Command R+ 104B}} & Og & \textbf{79.29} & \textbf{70.34} & \textbf{83.20} & \textbf{79.08} & 76.26 & 74.54 & 64.98 & 58.66 & 65.90 & 61.36 & 73.93 & 68.80 \\
 & En & \underline{80.00} & \underline{77.28} & \underline{81.92} & \underline{75.50} & 80.40 & 75.40 & 66.43 & 58.78 & - & - & 77.19 & 71.74 \\
\multirow[t]{2}{*}{\texttt{Qwen 2.5 72B Instruct}} & Og & 70.87 & 69.69 & 66.30 & 60.77 & 81.68 & 77.87 & 74.78 & 67.59 & 72.56 & 67.04 & 73.24 & 68.59 \\
 & En & 76.16 & 68.48 & 69.70 & 61.76 & \underline{82.00} & 77.52 & 76.55 & 68.46 & - & - & 76.10 & 69.06 \\
\multirow[t]{2}{*}{\texttt{Llama 3.1 70B Instruct}} & Og & 69.99 & 62.54 & 72.08 & 67.69 & 60.64 & 61.20 & 75.90 & 68.65 & 77.86 & 72.07 & 71.29 & 66.43 \\
 & En & 71.87 & 63.77 & 75.27 & 67.46 & 77.01 & 76.79 & 78.29 & 70.66 & - & - & 75.61 & 69.67 \\ \hline
 \multirow[t]{2}{*}{Average} & Og & 73.05 & 67.42 & 73.95 & 69.72 & 75.26 & 73.54 & 73.77 & 66.63 & 74.58 & 68.69 & 74.12 & 69.20 \\
 & En & 75.88 & 69.24 & 76.57 & 69.28 & 80.26 & 76.94 & 75.06 & 67.50 & - & - & 76.94 & 70.74 \\
\midrule
\multicolumn{14}{l}{\quad \textit{LLMs with less than 10B parameters (10B- LLMs)}} \\
\midrule 
\multirow[t]{2}{*}{\texttt{Llama 3.1 8B Instruct}} & Og & 48.63 & 47.19 & 70.88 & 69.70 & 47.78 & 42.15 & 68.71 & 62.61 & \textbf{71.77} & 64.48 & 61.55 & 57.23 \\
 & En & 60.57 & 53.51 & 76.48 & 71.11 & 64.22 & 56.93 & 74.27 & 66.33 & - & - & 68.89 & 61.97 \\
\multirow[t]{2}{*}{\texttt{Qwen 2.5 7B Instruct}} & Og & \textbf{65.40} & \textbf{64.18} & \textbf{79.95} & \textbf{74.44} & \textbf{71.43} & \textbf{72.52} & \textbf{80.74} & \textbf{75.15} & 66.97 & \textbf{66.49} & \textbf{72.90} & \textbf{70.56} \\
 & En & 66.76 & \underline{64.02} & \underline{81.07} & \underline{76.09} & \underline{69.02} & \underline{70.89} & \underline{81.02} & \underline{75.10} & - & - & \underline{74.47} & \underline{71.53} \\
\multirow[t]{2}{*}{\texttt{Mistral v3 7B}} & Og & 64.92 & 61.26 & 65.98 & 60.46 & 57.91 & 53.31 & 72.97 & 67.36 & 65.67 & 59.10 & 65.49 & 60.30 \\
 & En & \underline{68.69} & 63.94 & 71.01 & 63.49 & 68.67 & 62.79 & 74.00 & 66.08 & - & - & 70.59 & 64.08 \\ \hline
 \multirow[t]{2}{*}{Average} & Og & 59.65 & 57.54 & 72.27 & 68.20 & 59.04 & 55.99 & 74.14 & 68.37 & 68.14 & 63.36 & 66.65 & 62.69 \\
 & En & 65.34 & 60.49 & 76.19 & 70.23 & 67.30 & 63.54 & 76.43 & 69.17 & - & - & 71.32 & 65.86 \\
\bottomrule
\end{tabular*}
\caption{Performance comparison of LLMs and baselines in monolingual and cross-lingual settings using Macro F1 score. The best results for the version with original language (Og) are in \textbf{bold}, and for the version with English translations (En) are \underline{underlined} for each category. 
}
\label{tab:cross-lingual-comparison-macro}
\end{table*}

\subsection{Monolingual Evaluation}
\label{sec:mono}

Figure~\ref{fig:large_models_languages} (left-hand side) highlights the average performance across prompting techniques for each LLM. The capabilities to process languages vary among LLMs. For example, \texttt{C4AI Command R+}, trained on 23 languages, exhibits high performance across many languages. However, even \textbf{LLMs that cover fewer languages can perform well} (e.g., \texttt{Mistral Large}). This suggests that 70B+ LLMs demonstrate generalization on multilingual data.

In monolingual settings, some languages perform poorly, which is mostly the case for Slavic languages, Hungarian and Burmese, where the performance was lower than for other languages and language families. In contrast, \textbf{high-resource languages achieved superior results in most cases} ($p<0.05$). However, these results depend not only on the language but also on the data's complexity -- specifically, variations in data across languages and other attributes that affect how easily the LLM can predict the correct answer. This was evident since performance differences persisted even after translating all data into English. Additionally, factors such as topic distribution may influence performance, with languages that cover a wider range of topics potentially benefiting from this.

\texttt{DeBERTa v3 Large}, fine-tuned on NLI data, performed well in monolingual settings for many languages, often outperforming weaker LLMs. However, it struggled with low-resource and non-Latin languages, resulting in lower average performance overall. Other baselines also underperformed, highlighting the superior generalization ability of LLMs in identifying claim-post relevance.



\subsection{Cross-Lingual Evaluation}
\label{sec:cross}

We compare LLMs performance in monolingual and cross-lingual settings with Macro F1 scores presented in Table~\ref{tab:cross-lingual-comparison-macro} and across techniques for each language combination in Figure~\ref{fig:large_models_languages} (right-hand side). Overall, \textbf{performance declined in cross-lingual settings}, with an average decrease of approximately 4.5\%, especially for 10B- LLMs. This highlights the challenges of processing inputs in different languages. 

Few-shot prompting proved effective when applied to multilingual LLMs. In contrast, \textbf{CoT helped mitigate the cross-lingual performance gap for smaller models} (5\% improvement compared to XLT prompting), providing a reasoning approach that transfers well across languages.

\textbf{The \texttt{C4AI Command R+} model exhibited superior performance for a monolingual scenario} in zero-shot settings ($p<0.05$). In contrast, \texttt{Mistral Large} emerged as the best-performing LLM in few-shot settings and reasoning (CoT and XLT prompting) with original language inputs ($p<0.05$). These findings suggest that LLMs with less extensive language coverage during training can outperform highly multilingual LLMs when advanced prompting techniques are leveraged. 

For 10B- LLMs, \textbf{\texttt{Qwen2.5 7B} consistently achieved superior performance in both settings} across prompting techniques ($p<0.05$), excluding XLT in monolingual settings. This demonstrates the effectiveness of \texttt{Qwen2.5}'s training in equipping the LLM with strong generalization capabilities across different prompting strategies.

\subsection{Translation-Based Approaches}
\label{sec:english}

\begin{figure}
\centering
\vspace{-2mm}
\includegraphics[width=\columnwidth]{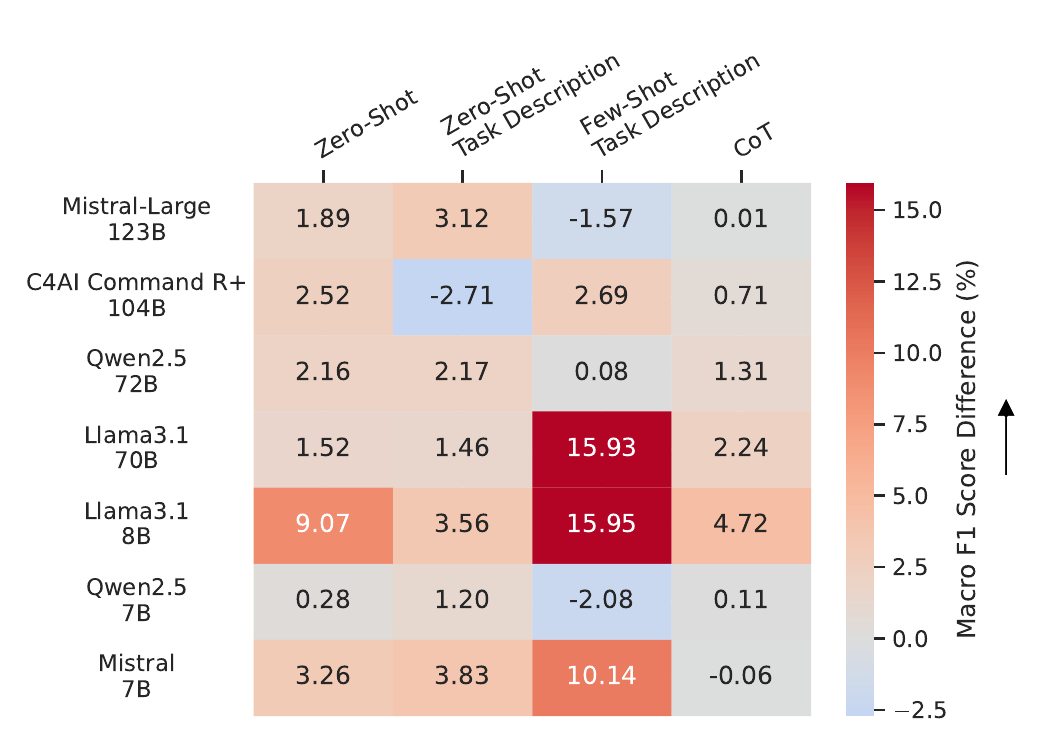}
\caption{Overall difference between English translation and the original language using Macro F1 score.}
\label{fig:eng-vs-org}
\end{figure}

We analyze the performance difference between original language inputs and their English translations, as illustrated in Figure~\ref{fig:eng-vs-org}. The results reveal that \textbf{English translation generally enhanced LLM performance across most scenarios} ($p<0.05$). This finding highlights the potential of translation-based approaches (translating to English or using XLT) for enhancing the performance of models with limited multilingual capabilities.

English translations not only improve cross-lingual performance but also demonstrate that LLMs often achieve higher accuracy when operating in English. However, \textbf{English translations can sometimes negatively impact performance}, as observed for the \texttt{C4AI Command R+} model. This LLM, trained with a higher number of languages, performed better with original language inputs, suggesting that extensive multilingual training may outperform translation-based strategies.

\textbf{LLMs trained predominantly on Latin-script data}, such as \texttt{Llama3.1}, \textbf{showed significant performance gains when translations to English} were employed (e.g., a 16\% improvement in Macro F1 in few-shot settings), observed as statistically significant ($p<0.05$). Translation-based approaches also proved effective in addressing non-Latin scripts ($p<0.05$), making them a practical alternative in cross-lingual settings. 

However, across almost all LLMs, \textbf{CoT prompting combined with English translations yielded only marginal improvements} ($p<0.05$), suggesting that CoT prompting alone can serve as a viable substitute for translation in achieving comparable performance. This highlights the potential of reasoning-based strategies to bridge cross-lingual gaps without relying on intermediate translations. Additionally, the impact of English translations proved less effective for few-shot settings, with \texttt{Mistral Large} and \texttt{Qwen2.5 7B} models showing a negative effect on the results in these scenarios.

\section{Error Analysis}

This section analyzes the errors in the reasoning generated by LLMs, focusing on the CoT and XLT techniques. We categorize errors into two types. First, \textit{output consistency} errors, where LLM's responses are inconsistent or deviate from the expected format. Second, \textit{reasoning} errors, which account for misclassified relevance pairs.

\subsection{Output Consistency Errors}

We used automatic tools to detect output consistency errors, including FastText~\cite{joulin2016bag, joulin2016fasttext} and langdetect for language identification, and sequence occurrence analysis for detecting repetitions (see Appendix~\ref{app:error-analysis}).


A commonly observed issue relates to the \textbf{language of LLM outputs}, as all responses were expected in English. \texttt{Llama3.1} models with CoT prompting frequently generated non-English outputs -- 37\% for 8B version and 28\% for 70B version. More details are given in Table~\ref{tab:error-analysis} in Appendix~\ref{app:error-analysis}.

Other errors included \textbf{repeating sequences}, mainly observed in \texttt{Llama3.1 8B} with CoT prompting (215$\times$). Other LLMs, such as \texttt{Llama3.1} (33$\times$), \texttt{Mistral Large} (16$\times$) and \texttt{Qwen2.5 72B} (11$\times$), exhibited fewer occurrences using XLT prompting. \texttt{Llama3.1} refused to generate responses for five pairs due to the disinformation content.

\subsection{Reasoning Errors}

We selected 20 random samples incorrectly classified for each LLM and for the CoT and XLT techniques, which we manually reviewed to identify common reasoning errors. For CoT, we also included the version with English translations, which resulted in 420 annotated samples overall. Our analysis revealed several types of errors, particularly those contributing to false positives. 

The most common error was \textbf{incorrect reasoning based on topic similarity} (around 65\%), where posts and fact-checked claims were misclassified as relevant based solely on shared topics. This was especially frequent with COVID-19, vaccination topics and cases when both statements are attributed to the same entity. Some incorrectly identified samples exhibit contradictory reasoning (approximately 7\%), mostly for \texttt{Mistral 7B} with CoT. For example, while individual statements are correctly classified as irrelevant, the LLM focused on the topic similarity rather than the actual irrelevance, leading to misclassifications (see Appendix~\ref{app:cosistency-errors}).

Other reasoning errors arise from \textbf{missing context} in posts or fact-checked claims, especially when referencing images or videos that LLMs cannot process or that are irrelevant. We thus ignored posts that contained visual information through links or embedded media during the selection process for our annotation. Some of the posts were missing URL links, but referred to images. Such posts with visual information were not filtered.

\section{Discussion}

\paragraph{Multilingual Previously Fact-Checked Claim Detection and Low-Resource Languages.}

Our experiments revealed that LLMs work reasonably well in English and high-resource languages, demonstrating robust capabilities in detecting previously fact-checked claims. However, a notable performance gap persists for some low-resource languages and those with non-Latin scripts. This disparity emphasizes the need for tailored adaptations, particularly for non-English settings.



\paragraph{Superiority of Translations-Based Approaches.} 

Translation-based approaches were particularly effective for low-resource languages and non-Latin scripts, as well as when using 10B- LLMs. Translating inputs into English (using machine translation) allows LLMs to benefit from their extensive training in English, which typically provides more robust results. This method is useful in scenarios where processing of low-resource languages would otherwise lead to suboptimal outcomes.

\paragraph{Prompting Techniques.}

No single prompting technique emerged as universally superior across all settings. Zero-shot was beneficial for high-resource languages, but did not work well with low-resource languages due to limited contextual understanding and sparse pre-training data. Few-shot prompting showed improvements in low-resource languages, but required carefully selected samples.

For high-resource scenarios, using larger LLMs with few-shot prompting in the original language provides reliable results across languages. In contrast, resource-constrained scenarios benefited from combining 10B- LLMs, CoT and translation-based approaches. These findings emphasize that the choice of technique should be guided by specific languages and other considerations.

\section{Conclusion}

This paper presents a comprehensive evaluation of seven LLMs, ranging from 7B up to 123B parameters, for detecting previously fact-checked claims in monolingual and cross-lingual settings. We created and released a new dataset consisting of 16,000 manually annotated post-claim pairs across 20 languages. Our findings show that bigger LLMs perform best overall, while smaller models like \texttt{Qwen2.5 7B} can compete effectively when combined with CoT prompting.

We observed consistent gaps in performance for low-resource languages and non-Latin scripts. In practical settings, this suggests that translation-based approaches remain necessary to ensure robust results across languages. Our findings underscore the importance of both LLM selection and prompting strategies in optimizing performance for previously fact-checked claim detection tasks.

\section*{Limitations}

\paragraph{Model Selection.}

Our study focused on state-of-the-art LLMS that are openly available. We excluded closed-source models like GPT-4 since our experiments required analyzing token probabilities, which are only accessible in open-source models. Additionally, open-source models offer greater experimental control compared to closed-source LLMs. Furthermore, our analysis considered models released before July 2024, which marked the primary research period of this study.

\paragraph{Language Support.}

The selected LLMs exhibit varying degrees of multilingual capabilities, ranging from primarily English-centric models to those supporting 29 languages. While model cards indicate intended language support, the models may demonstrate capabilities in additional languages due to the training data diversity and potential data contamination. Our analysis spans 20 languages across different language families and writing systems, making multilingual support a key selection criterion. Although some languages in our study lack explicit support in any of the evaluated models, we assume that the models might still demonstrate some capacity to assess text similarity in these languages. This setup enabled us to evaluate the models' multilingual capabilities and compare their performance for PFCD across different languages.



\paragraph{Language Detection for Error Analysis.}

For error analysis of LLM outputs, we employed language identification using two tools, especially FastText and langdetect. Due to the varying accuracy across different languages (mostly concerned with low-resource languages), we employed both tools in parallel for the final language analysis. Outputs were identified as a different language than English when both tools agreed on identifying a non-English language, providing a more robust detection mechanism for language-related errors in model responses. However, the performance of these tools can vary across languages, and their performance can be lower for low-resource languages, which can result in incorrect identification of the language for some inputs.

\section*{Ethical Consideration}

\paragraph{Intended Use.} The annotated dataset is intended primarily for research purposes and is derived from the existing \textbf{\textit{MultiClaim}} dataset~\cite{pikuliak-etal-2023-multilingual}. In our work, we selected a subset of \textbf{\textit{MultiClaim}} and annotated a portion of the data, specifically assessing the relevance between social media posts and fact-checked claims. Along with the dataset, we also release code to reproduce our results. Both the datasets and code are only for research use, and reproducing the results requires access to the original \textbf{\textit{MultiClaim}} dataset.

\paragraph{Usage of AI Assistants.} We have used the AI assistant for grammar checks and sentence structure improvements. We have not used AI assistants in the research process beyond the experiments detailed in the Methodology section (Sec. \ref{sec:methodology}). 

\section*{Acknowledgments}

This research was partially supported by \textit{DisAI - Improving scientific excellence and creativity in combating disinformation with artificial intelligence and language technologies}, a project funded by Horizon Europe under \href{https://doi.org/10.3030/101079164}{GA No.101079164}, by \textit{MODERMED}, a project funded by the Slovak Research and Development Agency under GA No. APVV-22-0414, and by the \textit{European Union NextGenerationEU} through the Recovery and Resilience Plan for Slovakia under the project No. 09I01-03-V04-00007.

This work was supported by the Ministry of Education, Youth and Sports of the Czech Republic through the e-INFRA CZ (ID:90254).

\bibliography{custom}

\begin{thebibliography}{35}
\providecommand{\natexlab}[1]{#1}

\bibitem[{Agresti and Coull(1998)}]{Agresti01051998}
Alan Agresti and Brent~A. Coull. 1998.
\newblock \href {https://doi.org/10.1080/00031305.1998.10480550} {Approximate is better than “exact” for interval estimation of binomial proportions}.
\newblock \emph{The American Statistician}, 52(2):119--126.

\bibitem[{A{\"\i}meur et~al.(2023)A{\"\i}meur, Amri, and Brassard}]{aimeur2023fake}
Esma A{\"\i}meur, Sabrine Amri, and Gilles Brassard. 2023.
\newblock Fake news, disinformation and misinformation in social media: a review.
\newblock \emph{Social Network Analysis and Mining}, 13(1):30.

\bibitem[{Barr{\'o}n-Cede{\~{n}}o et~al.(2020)Barr{\'o}n-Cede{\~{n}}o, Elsayed, Nakov, Da~San~Martino, Hasanain, Suwaileh, Haouari, Babulkov, Hamdan, Nikolov, Shaar, and Ali}]{barroncedeno2020overviewcheckthat2020automatic}
Alberto Barr{\'o}n-Cede{\~{n}}o, Tamer Elsayed, Preslav Nakov, Giovanni Da~San~Martino, Maram Hasanain, Reem Suwaileh, Fatima Haouari, Nikolay Babulkov, Bayan Hamdan, Alex Nikolov, Shaden Shaar, and Zien~Sheikh Ali. 2020.
\newblock Overview of checkthat! 2020: Automatic identification and verification of claims in social media.
\newblock In \emph{Experimental IR Meets Multilinguality, Multimodality, and Interaction}, pages 215--236, Cham. Springer International Publishing.

\bibitem[{Brown et~al.(2020)Brown, Mann, Ryder, Subbiah, Kaplan, Dhariwal, Neelakantan, Shyam, Sastry, Askell, Agarwal, Herbert-Voss, Krueger, Henighan, Child, Ramesh, Ziegler, Wu, Winter, Hesse, Chen, Sigler, Litwin, Gray, Chess, Clark, Berner, McCandlish, Radford, Sutskever, and Amodei}]{NEURIPS2020_1457c0d6}
Tom Brown, Benjamin Mann, Nick Ryder, Melanie Subbiah, Jared~D Kaplan, Prafulla Dhariwal, Arvind Neelakantan, Pranav Shyam, Girish Sastry, Amanda Askell, Sandhini Agarwal, Ariel Herbert-Voss, Gretchen Krueger, Tom Henighan, Rewon Child, Aditya Ramesh, Daniel Ziegler, Jeffrey Wu, Clemens Winter, and 12 others. 2020.
\newblock \href {https://proceedings.neurips.cc/paper_files/paper/2020/file/1457c0d6bfcb4967418bfb8ac142f64a-Paper.pdf} {Language models are few-shot learners}.
\newblock In \emph{Advances in Neural Information Processing Systems}, volume~33, pages 1877--1901. Curran Associates, Inc.

\bibitem[{Choi and Ferrara(2024{\natexlab{a}})}]{10.1145/3589335.3651910}
Eun~Cheol Choi and Emilio Ferrara. 2024{\natexlab{a}}.
\newblock \href {https://doi.org/10.1145/3589335.3651910} {Automated claim matching with large language models: Empowering fact-checkers in the fight against misinformation}.
\newblock In \emph{Companion Proceedings of the ACM Web Conference 2024}, WWW '24, page 1441–1449, New York, NY, USA. Association for Computing Machinery.

\bibitem[{Choi and Ferrara(2024{\natexlab{b}})}]{10.1145/3589335.3651504}
Eun~Cheol Choi and Emilio Ferrara. 2024{\natexlab{b}}.
\newblock \href {https://doi.org/10.1145/3589335.3651504} {Fact-gpt: Fact-checking augmentation via claim matching with llms}.
\newblock In \emph{Companion Proceedings of the ACM Web Conference 2024}, WWW '24, page 883–886, New York, NY, USA. Association for Computing Machinery.

\bibitem[{{Cohere For AI}(2024)}]{cohere_for_ai_2024}
{Cohere For AI}. 2024.
\newblock \href {https://doi.org/10.57967/hf/3136} {c4ai-command-r-plus-08-2024 (revision dfda5ab)}.

\bibitem[{Grattafiori et~al.(2024)Grattafiori, Dubey, Jauhri, Pandey, Kadian, Al-Dahle, Letman, Mathur, Schelten, Vaughan, Yang, Fan, Goyal, Hartshorn, Yang, Mitra, Sravankumar, Korenev, Hinsvark, Rao, Zhang, Rodriguez, Gregerson, Spataru, Roziere, Biron, Tang, Chern, Caucheteux, Nayak, Bi, Marra, McConnell, Keller, Touret, Wu, Wong, Ferrer, Nikolaidis, Allonsius, Song, Pintz, Livshits, Wyatt, Esiobu, Choudhary, Mahajan, Garcia-Olano, Perino, Hupkes, Lakomkin, AlBadawy, Lobanova, Dinan, Smith, Radenovic, Guzmán, Zhang, Synnaeve, Lee, Anderson, Thattai, Nail, Mialon, Pang, Cucurell, Nguyen, Korevaar, Xu, Touvron, Zarov, Ibarra, Kloumann, Misra, Evtimov, Zhang, Copet, Lee, Geffert, Vranes, Park, Mahadeokar, Shah, van~der Linde, Billock, Hong, Lee, Fu, Chi, Huang, Liu, Wang, Yu, Bitton, Spisak, Park, Rocca, Johnstun, Saxe, Jia, Alwala, Prasad, Upasani, Plawiak, Li, Heafield, Stone, El-Arini, Iyer, Malik, Chiu, Bhalla, Lakhotia, Rantala-Yeary, van~der Maaten, Chen, Tan, Jenkins, Martin, Madaan, Malo, Blecher,
  Landzaat, de~Oliveira, Muzzi, Pasupuleti, Singh, Paluri, Kardas, Tsimpoukelli, Oldham, Rita, Pavlova, Kambadur, Lewis, Si, Singh, Hassan, Goyal, Torabi, Bashlykov, Bogoychev, Chatterji, Zhang, Duchenne, Çelebi, Alrassy, Zhang, Li, Vasic, Weng, Bhargava, Dubal, Krishnan, Koura, Xu, He, Dong, Srinivasan, Ganapathy, Calderer, Cabral, Stojnic, Raileanu, Maheswari, Girdhar, Patel, Sauvestre, Polidoro, Sumbaly, Taylor, Silva, Hou, Wang, Hosseini, Chennabasappa, Singh, Bell, Kim, Edunov, Nie, Narang, Raparthy, Shen, Wan, Bhosale, Zhang, Vandenhende, Batra, Whitman, Sootla, Collot, Gururangan, Borodinsky, Herman, Fowler, Sheasha, Georgiou, Scialom, Speckbacher, Mihaylov, Xiao, Karn, Goswami, Gupta, Ramanathan, Kerkez, Gonguet, Do, Vogeti, Albiero, Petrovic, Chu, Xiong, Fu, Meers, Martinet, Wang, Wang, Tan, Xia, Xie, Jia, Wang, Goldschlag, Gaur, Babaei, Wen, Song, Zhang, Li, Mao, Coudert, Yan, Chen, Papakipos, Singh, Srivastava, Jain, Kelsey, Shajnfeld, Gangidi, Victoria, Goldstand, Menon, Sharma, Boesenberg,
  Baevski, Feinstein, Kallet, Sangani, Teo, Yunus, Lupu, Alvarado, Caples, Gu, Ho, Poulton, Ryan, Ramchandani, Dong, Franco, Goyal, Saraf, Chowdhury, Gabriel, Bharambe, Eisenman, Yazdan, James, Maurer, Leonhardi, Huang, Loyd, Paola, Paranjape, Liu, Wu, Ni, Hancock, Wasti, Spence, Stojkovic, Gamido, Montalvo, Parker, Burton, Mejia, Liu, Wang, Kim, Zhou, Hu, Chu, Cai, Tindal, Feichtenhofer, Gao, Civin, Beaty, Kreymer, Li, Adkins, Xu, Testuggine, David, Parikh, Liskovich, Foss, Wang, Le, Holland, Dowling, Jamil, Montgomery, Presani, Hahn, Wood, Le, Brinkman, Arcaute, Dunbar, Smothers, Sun, Kreuk, Tian, Kokkinos, Ozgenel, Caggioni, Kanayet, Seide, Florez, Schwarz, Badeer, Swee, Halpern, Herman, Sizov, Guangyi, Zhang, Lakshminarayanan, Inan, Shojanazeri, Zou, Wang, Zha, Habeeb, Rudolph, Suk, Aspegren, Goldman, Zhan, Damlaj, Molybog, Tufanov, Leontiadis, Veliche, Gat, Weissman, Geboski, Kohli, Lam, Asher, Gaya, Marcus, Tang, Chan, Zhen, Reizenstein, Teboul, Zhong, Jin, Yang, Cummings, Carvill, Shepard, McPhie,
  Torres, Ginsburg, Wang, Wu, U, Saxena, Khandelwal, Zand, Matosich, Veeraraghavan, Michelena, Li, Jagadeesh, Huang, Chawla, Huang, Chen, Garg, A, Silva, Bell, Zhang, Guo, Yu, Moshkovich, Wehrstedt, Khabsa, Avalani, Bhatt, Mankus, Hasson, Lennie, Reso, Groshev, Naumov, Lathi, Keneally, Liu, Seltzer, Valko, Restrepo, Patel, Vyatskov, Samvelyan, Clark, Macey, Wang, Hermoso, Metanat, Rastegari, Bansal, Santhanam, Parks, White, Bawa, Singhal, Egebo, Usunier, Mehta, Laptev, Dong, Cheng, Chernoguz, Hart, Salpekar, Kalinli, Kent, Parekh, Saab, Balaji, Rittner, Bontrager, Roux, Dollar, Zvyagina, Ratanchandani, Yuvraj, Liang, Alao, Rodriguez, Ayub, Murthy, Nayani, Mitra, Parthasarathy, Li, Hogan, Battey, Wang, Howes, Rinott, Mehta, Siby, Bondu, Datta, Chugh, Hunt, Dhillon, Sidorov, Pan, Mahajan, Verma, Yamamoto, Ramaswamy, Lindsay, Lindsay, Feng, Lin, Zha, Patil, Shankar, Zhang, Zhang, Wang, Agarwal, Sajuyigbe, Chintala, Max, Chen, Kehoe, Satterfield, Govindaprasad, Gupta, Deng, Cho, Virk, Subramanian, Choudhury,
  Goldman, Remez, Glaser, Best, Koehler, Robinson, Li, Zhang, Matthews, Chou, Shaked, Vontimitta, Ajayi, Montanez, Mohan, Kumar, Mangla, Ionescu, Poenaru, Mihailescu, Ivanov, Li, Wang, Jiang, Bouaziz, Constable, Tang, Wu, Wang, Wu, Gao, Kleinman, Chen, Hu, Jia, Qi, Li, Zhang, Zhang, Adi, Nam, Yu, Wang, Zhao, Hao, Qian, Li, He, Rait, DeVito, Rosnbrick, Wen, Yang, Zhao, and Ma}]{grattafiori2024llama3herdmodels}
Aaron Grattafiori, Abhimanyu Dubey, Abhinav Jauhri, Abhinav Pandey, Abhishek Kadian, Ahmad Al-Dahle, Aiesha Letman, Akhil Mathur, Alan Schelten, Alex Vaughan, Amy Yang, Angela Fan, Anirudh Goyal, Anthony Hartshorn, Aobo Yang, Archi Mitra, Archie Sravankumar, Artem Korenev, Arthur Hinsvark, and 542 others. 2024.
\newblock \href {https://arxiv.org/abs/2407.21783} {The llama 3 herd of models}.
\newblock \emph{Preprint}, arXiv:2407.21783.

\bibitem[{Hardalov et~al.(2022)Hardalov, Chernyavskiy, Koychev, Ilvovsky, and Nakov}]{hardalov2022crowdcheckeddetectingpreviouslyfactchecked}
Momchil Hardalov, Anton Chernyavskiy, Ivan Koychev, Dmitry Ilvovsky, and Preslav Nakov. 2022.
\newblock \href {https://doi.org/10.18653/v1/2022.aacl-main.22} {{C}rowd{C}hecked: Detecting previously fact-checked claims in social media}.
\newblock In \emph{Proceedings of the 2nd Conference of the Asia-Pacific Chapter of the Association for Computational Linguistics and the 12th International Joint Conference on Natural Language Processing (Volume 1: Long Papers)}, pages 266--285, Online only. Association for Computational Linguistics.

\bibitem[{Hrckova et~al.(2025)Hrckova, Moro, Srba, Simko, and Bielikova}]{hrckova2024autonomationautomationactivitiesneeds}
Andrea Hrckova, Robert Moro, Ivan Srba, Jakub Simko, and Maria Bielikova. 2025.
\newblock \href {https://doi.org/10.1145/3764592} {Autonomation, not automation: Activities and needs of european fact-checkers as a basis for designing human-centered ai systems}.
\newblock \emph{ACM J. Responsib. Comput.}
\newblock Just Accepted.

\bibitem[{Huang et~al.(2023)Huang, Tang, Zhang, Zhao, Song, Xia, and Wei}]{huang-etal-2023-languages}
Haoyang Huang, Tianyi Tang, Dongdong Zhang, Xin Zhao, Ting Song, Yan Xia, and Furu Wei. 2023.
\newblock \href {https://doi.org/10.18653/v1/2023.findings-emnlp.826} {Not all languages are created equal in {LLM}s: Improving multilingual capability by cross-lingual-thought prompting}.
\newblock In \emph{Findings of the Association for Computational Linguistics: EMNLP 2023}, pages 12365--12394, Singapore. Association for Computational Linguistics.

\bibitem[{Jiang et~al.(2023)Jiang, Sablayrolles, Mensch, Bamford, Chaplot, de~las Casas, Bressand, Lengyel, Lample, Saulnier, Lavaud, Lachaux, Stock, Scao, Lavril, Wang, Lacroix, and Sayed}]{jiang2023mistral7b}
Albert~Q. Jiang, Alexandre Sablayrolles, Arthur Mensch, Chris Bamford, Devendra~Singh Chaplot, Diego de~las Casas, Florian Bressand, Gianna Lengyel, Guillaume Lample, Lucile Saulnier, Lélio~Renard Lavaud, Marie-Anne Lachaux, Pierre Stock, Teven~Le Scao, Thibaut Lavril, Thomas Wang, Timothée Lacroix, and William~El Sayed. 2023.
\newblock \href {https://arxiv.org/abs/2310.06825} {Mistral 7b}.
\newblock \emph{Preprint}, arXiv:2310.06825.

\bibitem[{Joulin et~al.(2016)Joulin, Grave, Bojanowski, Douze, J{\'e}gou, and Mikolov}]{joulin2016fasttext}
Armand Joulin, Edouard Grave, Piotr Bojanowski, Matthijs Douze, H{\'e}rve J{\'e}gou, and Tomas Mikolov. 2016.
\newblock Fasttext.zip: Compressing text classification models.
\newblock \emph{arXiv preprint arXiv:1612.03651}.

\bibitem[{Joulin et~al.(2017)Joulin, Grave, Bojanowski, and Mikolov}]{joulin2016bag}
Armand Joulin, Edouard Grave, Piotr Bojanowski, and Tomas Mikolov. 2017.
\newblock \href {https://aclanthology.org/E17-2068/} {Bag of tricks for efficient text classification}.
\newblock In \emph{Proceedings of the 15th Conference of the {E}uropean Chapter of the Association for Computational Linguistics: Volume 2, Short Papers}, pages 427--431, Valencia, Spain. Association for Computational Linguistics.

\bibitem[{Kazemi et~al.(2021)Kazemi, Garimella, Gaffney, and Hale}]{kazemi2021claimmatchingenglishscale}
Ashkan Kazemi, Kiran Garimella, Devin Gaffney, and Scott Hale. 2021.
\newblock \href {https://doi.org/10.18653/v1/2021.acl-long.347} {Claim matching beyond {E}nglish to scale global fact-checking}.
\newblock In \emph{Proceedings of the 59th Annual Meeting of the Association for Computational Linguistics and the 11th International Joint Conference on Natural Language Processing (Volume 1: Long Papers)}, pages 4504--4517, Online. Association for Computational Linguistics.

\bibitem[{Kazemi et~al.(2022)Kazemi, Li, Pérez-Rosas, Hale, and Mihalcea}]{kazemi2022matchingtweetsapplicablefactchecks}
Ashkan Kazemi, Zehua Li, Verónica Pérez-Rosas, Scott~A. Hale, and Rada Mihalcea. 2022.
\newblock \href {https://arxiv.org/abs/2202.07094} {Matching tweets with applicable fact-checks across languages}.
\newblock \emph{Preprint}, arXiv:2202.07094.

\bibitem[{Kruskal and Wallis(1952)}]{Kruskal01121952}
William~H. Kruskal and W.~Allen Wallis. 1952.
\newblock \href {https://doi.org/10.1080/01621459.1952.10483441} {Use of ranks in one-criterion variance analysis}.
\newblock \emph{Journal of the American Statistical Association}, 47(260):583--621.

\bibitem[{Larraz et~al.(2023)Larraz, M{\'\i}guez, and Sallicati}]{larraz2023semantic}
Irene Larraz, Rub{\'e}n M{\'\i}guez, and Francesca Sallicati. 2023.
\newblock Semantic similarity models for automated fact-checking: Claimcheck as a claim matching tool.
\newblock \emph{Profesional de la Informaci{\'o}n}, 32(3).

\bibitem[{Mann and Whitney(1947)}]{497e1044-d5b0-30a9-b230-3ca0f10d6f6c}
H.~B. Mann and D.~R. Whitney. 1947.
\newblock \href {http://www.jstor.org/stable/2236101} {On a test of whether one of two random variables is stochastically larger than the other}.
\newblock \emph{The Annals of Mathematical Statistics}, 18(1):50--60.

\bibitem[{Micallef et~al.(2022)Micallef, Armacost, Memon, and Patil}]{10.1145/3512974}
Nicholas Micallef, Vivienne Armacost, Nasir Memon, and Sameer Patil. 2022.
\newblock \href {https://doi.org/10.1145/3512974} {True or false: Studying the work practices of professional fact-checkers}.
\newblock \emph{Proc. ACM Hum.-Comput. Interact.}, 6(CSCW1).

\bibitem[{{Mistral AI Team}(2024)}]{mistral-large}
{Mistral AI Team}. 2024.
\newblock \href {https://mistral.ai/news/mistral-large-2407/} {Large enough}.

\bibitem[{Nakov et~al.(2021)Nakov, Corney, Hasanain, Alam, Elsayed, Barrón-Cedeño, Papotti, Shaar, and Da~San~Martino}]{nakov2021automatedfactcheckingassistinghuman}
Preslav Nakov, David Corney, Maram Hasanain, Firoj Alam, Tamer Elsayed, Alberto Barrón-Cedeño, Paolo Papotti, Shaden Shaar, and Giovanni Da~San~Martino. 2021.
\newblock \href {https://doi.org/10.24963/ijcai.2021/619} {Automated fact-checking for assisting human fact-checkers}.
\newblock In \emph{Proceedings of the Thirtieth International Joint Conference on Artificial Intelligence, {IJCAI-21}}, pages 4551--4558. International Joint Conferences on Artificial Intelligence Organization.
\newblock Survey Track.

\bibitem[{Neumann et~al.(2023)Neumann, Kolossa, and Nickel}]{neumann-etal-2023-deep}
Anna Neumann, Dorothea Kolossa, and Robert~M Nickel. 2023.
\newblock \href {https://aclanthology.org/2023.icnlsp-1.13} {Deep learning-based claim matching with multiple negatives training}.
\newblock In \emph{Proceedings of the 6th International Conference on Natural Language and Speech Processing (ICNLSP 2023)}, pages 134--139, Online. Association for Computational Linguistics.

\bibitem[{Pikuliak et~al.(2023)Pikuliak, Srba, Moro, Hromadka, Smole{\v{n}}, Meli{\v{s}}ek, Vykopal, Simko, Podrou{\v{z}}ek, and Bielikova}]{pikuliak-etal-2023-multilingual}
Mat{\'u}{\v{s}} Pikuliak, Ivan Srba, Robert Moro, Timo Hromadka, Timotej Smole{\v{n}}, Martin Meli{\v{s}}ek, Ivan Vykopal, Jakub Simko, Juraj Podrou{\v{z}}ek, and Maria Bielikova. 2023.
\newblock \href {https://doi.org/10.18653/v1/2023.emnlp-main.1027} {Multilingual previously fact-checked claim retrieval}.
\newblock In \emph{Proceedings of the 2023 Conference on Empirical Methods in Natural Language Processing}, pages 16477--16500, Singapore. Association for Computational Linguistics.

\bibitem[{Shaar et~al.(2020)Shaar, Babulkov, Da~San~Martino, and Nakov}]{shaar2020knownliedetectingpreviously}
Shaden Shaar, Nikolay Babulkov, Giovanni Da~San~Martino, and Preslav Nakov. 2020.
\newblock \href {https://doi.org/10.18653/v1/2020.acl-main.332} {That is a known lie: Detecting previously fact-checked claims}.
\newblock In \emph{Proceedings of the 58th Annual Meeting of the Association for Computational Linguistics}, pages 3607--3618, Online. Association for Computational Linguistics.

\bibitem[{Shaar et~al.(2022)Shaar, Georgiev, Alam, Da~San~Martino, Mohamed, and Nakov}]{shaar2022assistinghumanfactcheckersdetecting}
Shaden Shaar, Nikola Georgiev, Firoj Alam, Giovanni Da~San~Martino, Aisha Mohamed, and Preslav Nakov. 2022.
\newblock \href {https://doi.org/10.18653/v1/2022.findings-emnlp.151} {Assisting the human fact-checkers: Detecting all previously fact-checked claims in a document}.
\newblock In \emph{Findings of the Association for Computational Linguistics: EMNLP 2022}, pages 2069--2080, Abu Dhabi, United Arab Emirates. Association for Computational Linguistics.

\bibitem[{Shliselberg and Dori-Hacohen(2022)}]{shliselberg2022riet}
Michael Shliselberg and Shiri Dori-Hacohen. 2022.
\newblock Riet lab at checkthat!-2022: Improving decoder based re-ranking for claim matching.
\newblock In \emph{CLEF (Working Notes)}, pages 671--678.

\bibitem[{Singh et~al.(2024)Singh, Vargus, D{'}souza, Karlsson, Mahendiran, Ko, Shandilya, Patel, Mataciunas, O{'}Mahony, Zhang, Hettiarachchi, Wilson, Machado, Moura, Krzemi{\'n}ski, Fadaei, Ergun, Okoh, Alaagib, Mudannayake, Alyafeai, Chien, Ruder, Guthikonda, Alghamdi, Gehrmann, Muennighoff, Bartolo, Kreutzer, {\"U}st{\"u}n, Fadaee, and Hooker}]{singh-etal-2024-aya}
Shivalika Singh, Freddie Vargus, Daniel D{'}souza, B{\"o}rje Karlsson, Abinaya Mahendiran, Wei-Yin Ko, Herumb Shandilya, Jay Patel, Deividas Mataciunas, Laura O{'}Mahony, Mike Zhang, Ramith Hettiarachchi, Joseph Wilson, Marina Machado, Luisa Moura, Dominik Krzemi{\'n}ski, Hakimeh Fadaei, Irem Ergun, Ifeoma Okoh, and 14 others. 2024.
\newblock \href {https://doi.org/10.18653/v1/2024.acl-long.620} {Aya dataset: An open-access collection for multilingual instruction tuning}.
\newblock In \emph{Proceedings of the 62nd Annual Meeting of the Association for Computational Linguistics (Volume 1: Long Papers)}, pages 11521--11567, Bangkok, Thailand. Association for Computational Linguistics.

\bibitem[{Team et~al.(2025)Team, Kamath, Ferret, Pathak, Vieillard, Merhej, Perrin, Matejovicova, Ramé, Rivière, Rouillard, Mesnard, Cideron, bastien Grill, Ramos, Yvinec, Casbon, Pot, Penchev, Liu, Visin, Kenealy, Beyer, Zhai, Tsitsulin, Busa-Fekete, Feng, Sachdeva, Coleman, Gao, Mustafa, Barr, Parisotto, Tian, Eyal, Cherry, Peter, Sinopalnikov, Bhupatiraju, Agarwal, Kazemi, Malkin, Kumar, Vilar, Brusilovsky, Luo, Steiner, Friesen, Sharma, Sharma, Gilady, Goedeckemeyer, Saade, Feng, Kolesnikov, Bendebury, Abdagic, Vadi, György, Pinto, Das, Bapna, Miech, Yang, Paterson, Shenoy, Chakrabarti, Piot, Wu, Shahriari, Petrini, Chen, Lan, Choquette-Choo, Carey, Brick, Deutsch, Eisenbud, Cattle, Cheng, Paparas, Sreepathihalli, Reid, Tran, Zelle, Noland, Huizenga, Kharitonov, Liu, Amirkhanyan, Cameron, Hashemi, Klimczak-Plucińska, Singh, Mehta, Lehri, Hazimeh, Ballantyne, Szpektor, Nardini, Pouget-Abadie, Chan, Stanton, Wieting, Lai, Orbay, Fernandez, Newlan, yeong Ji, Singh, Black, Yu, Hui, Vodrahalli, Greff, Qiu,
  Valentine, Coelho, Ritter, Hoffman, Watson, Chaturvedi, Moynihan, Ma, Babar, Noy, Byrd, Roy, Momchev, Chauhan, Sachdeva, Bunyan, Botarda, Caron, Rubenstein, Culliton, Schmid, Sessa, Xu, Stanczyk, Tafti, Shivanna, Wu, Pan, Rokni, Willoughby, Vallu, Mullins, Jerome, Smoot, Girgin, Iqbal, Reddy, Sheth, Põder, Bhatnagar, Panyam, Eiger, Zhang, Liu, Yacovone, Liechty, Kalra, Evci, Misra, Roseberry, Feinberg, Kolesnikov, Han, Kwon, Chen, Chow, Zhu, Wei, Egyed, Cotruta, Giang, Kirk, Rao, Black, Babar, Lo, Moreira, Martins, Sanseviero, Gonzalez, Gleicher, Warkentin, Mirrokni, Senter, Collins, Barral, Ghahramani, Hadsell, Matias, Sculley, Petrov, Fiedel, Shazeer, Vinyals, Dean, Hassabis, Kavukcuoglu, Farabet, Buchatskaya, Alayrac, Anil, Dmitry, Lepikhin, Borgeaud, Bachem, Joulin, Andreev, Hardin, Dadashi, and Hussenot}]{gemmateam2025gemma3technicalreport}
Gemma Team, Aishwarya Kamath, Johan Ferret, Shreya Pathak, Nino Vieillard, Ramona Merhej, Sarah Perrin, Tatiana Matejovicova, Alexandre Ramé, Morgane Rivière, Louis Rouillard, Thomas Mesnard, Geoffrey Cideron, Jean bastien Grill, Sabela Ramos, Edouard Yvinec, Michelle Casbon, Etienne Pot, Ivo Penchev, and 197 others. 2025.
\newblock \href {https://arxiv.org/abs/2503.19786} {Gemma 3 technical report}.
\newblock \emph{Preprint}, arXiv:2503.19786.

\bibitem[{Vosoughi et~al.(2018)Vosoughi, Roy, and Aral}]{doi:10.1126/science.aap9559}
Soroush Vosoughi, Deb Roy, and Sinan Aral. 2018.
\newblock \href {https://doi.org/10.1126/science.aap9559} {The spread of true and false news online}.
\newblock \emph{Science}, 359(6380):1146--1151.

\bibitem[{Vykopal et~al.(2024)Vykopal, Pikuliak, Ostermann, and Šimko}]{vykopal2024generativelargelanguagemodels}
Ivan Vykopal, Matúš Pikuliak, Simon Ostermann, and Marián Šimko. 2024.
\newblock \href {https://arxiv.org/abs/2407.02351} {Generative large language models in automated fact-checking: A survey}.
\newblock \emph{Preprint}, arXiv:2407.02351.

\bibitem[{Wang et~al.(2024)Wang, Yang, Huang, Yang, Majumder, and Wei}]{wang2024multilinguale5textembeddings}
Liang Wang, Nan Yang, Xiaolong Huang, Linjun Yang, Rangan Majumder, and Furu Wei. 2024.
\newblock \href {https://arxiv.org/abs/2402.05672} {Multilingual e5 text embeddings: A technical report}.
\newblock \emph{Preprint}, arXiv:2402.05672.

\bibitem[{Wei et~al.(2024)Wei, Wang, Schuurmans, Bosma, Ichter, Xia, Chi, Le, and Zhou}]{10.5555/3600270.3602070}
Jason Wei, Xuezhi Wang, Dale Schuurmans, Maarten Bosma, Brian Ichter, Fei Xia, Ed~H. Chi, Quoc~V. Le, and Denny Zhou. 2024.
\newblock Chain-of-thought prompting elicits reasoning in large language models.
\newblock In \emph{Proceedings of the 36th International Conference on Neural Information Processing Systems}, NIPS '22, Red Hook, NY, USA. Curran Associates Inc.

\bibitem[{Yang et~al.(2024)Yang, Yang, Hui, Zheng, Yu, Zhou, Li, Li, Liu, Huang, Dong, Wei, Lin, Tang, Wang, Yang, Tu, Zhang, Ma, Yang, Xu, Zhou, Bai, He, Lin, Dang, Lu, Chen, Yang, Li, Xue, Ni, Zhang, Wang, Peng, Men, Gao, Lin, Wang, Bai, Tan, Zhu, Li, Liu, Ge, Deng, Zhou, Ren, Zhang, Wei, Ren, Liu, Fan, Yao, Zhang, Wan, Chu, Liu, Cui, Zhang, Guo, and Fan}]{yang2024qwen2technicalreport}
An~Yang, Baosong Yang, Binyuan Hui, Bo~Zheng, Bowen Yu, Chang Zhou, Chengpeng Li, Chengyuan Li, Dayiheng Liu, Fei Huang, Guanting Dong, Haoran Wei, Huan Lin, Jialong Tang, Jialin Wang, Jian Yang, Jianhong Tu, Jianwei Zhang, Jianxin Ma, and 43 others. 2024.
\newblock \href {https://arxiv.org/abs/2407.10671} {Qwen2 technical report}.
\newblock \emph{Preprint}, arXiv:2407.10671.

\bibitem[{Youden(1950)}]{youden-index}
W.~J. Youden. 1950.
\newblock \href {https://doi.org/10.1002/1097-0142(1950)3:1<32::AID-CNCR2820030106>3.0.CO;2-3} {Index for rating diagnostic tests}.
\newblock \emph{Cancer}, 3(1):32--35.

\end{thebibliography}

\appendix

\section{Computational Resources}

For our experiments, we leveraged a computational infrastructure consisting of A40 PCIe 40GB, A100 80GB and H100 NVL 94GB NVIDIA GPUs while our experiments ran in parallel on multiple GPUs. In total, our experiments required approximately 1500 GPU hours.

\section{Relevance Definition}
\label{sec:relevance}

Our definition of the \textit{relevance} differs from the textual entailment, since there is not only the strict relationship, whether a retrieved claim contradicts or entails with the given post. There is no general relation that always holds. Therefore, we defined the \textit{relevance} less strictly without considering the stance of the post or retrieved claim.

\paragraph{Theoretical Example.}

Given a claim \textit{"Vaccines cause autism"}, which can relate to claim \textit{"mRNA vaccines cause autism"}, which is more specific, but it also relates to claim \textit{"Vaccines are harmful"}, which is more general. In the first case, there is an entailment between the claims. However, in the second case, there is only partial overlap, while the entailment model can classify that as neutral. In the case of partial overlap, the fact-checker can reuse parts of the fact-checks and evidence that can be employed to verify the information.



\section{Human Annotation}
\label{app:human-annotation}

\subsection{Characteristics of Human Annotators} 

For the purpose of annotating a subset of \textit{\textbf{MultiClaim}} data, we employed six annotators. The annotators are all from our research team and have backgrounds in artificial intelligence and fact-checking. The annotation involved three men and three women, all from European countries, aged between 20 and 30 years.

\subsection{Annotation Process}

Each annotator was provided with both the original texts and their English translations, enabling them to annotate based on their proficiency in the original languages -- particularly if the annotator was a native speaker. However, due to the diversity of the 20 languages included in the study, in many cases, the annotators relied primarily on the English translations. Annotators were selected from various countries and were not limited to native speakers of a specific language. Each annotator was assigned a unique subset of data for annotation, and inter-annotator agreement was assessed using designed sets: \textit{pre-annotation} and \textit{post-annotation} sets, annotated before and after the annotation process.

During annotation, pairs were labeled as \textit{relevant}, \textit{irrelevant}, or \textit{cannot tell}. The \textit{cannot tell} label was used when annotators lacked sufficient context, encountered poor translations, or found the content unclear. To ensure full coverage, all \textit{cannot tell} classes were later re-annotated by one of the authors using additional context from the original \textbf{\textit{MultiClaim}} dataset, specifically the linked fact-checking articles. Based on additional context, each instance was classified as either \textit{relevant} or \textit{irrelevant}, resulting in a dataset with a binary classification.

\section{Analyzed Languages}

All languages and language pairs that are included in our experiments are listed in Table~\ref{tab:analyzed-languages} along with the proportion of relevant pairs annotated by human annotators.

Moreover, Table~\ref{tab:analyzed-languages-relevance} reports the proportion of relevant pairs identified for each language and language pair out of 400 pairs for each of them.

\section{Dataset}
\label{app:dataset}

Our dataset encompasses several popular topics, primarily related to COVID-19, the Russia-Ukraine war, vaccination, migration, and election fraud. Additionally, it includes the misattributed claims involving various politicians or public figures, such as Donald Trump, Greta Thunberg or George Orwell. Furthermore, the dataset covers region-specific topics that are prevalent in certain countries, such as claims related to Slovak politics or protests in specific regions.

In our work, we consider \textit{French}, \textit{Portuguese}, \textit{Spanish}, \textit{German}, \textit{Dutch}, \textit{English}, \textit{Arabic}, and \textit{Hindi} to be high-resource and other languages to be mid- or low-resource, based on~\citet{singh-etal-2024-aya}. We classified them based on the data available for training the LLMs.

\begin{table}[]
\resizebox{\columnwidth}{!}{%
\begin{tabular}{@{}llrrrr@{}}
\toprule
\textbf{Code} & \textbf{Language} & \multicolumn{1}{c}{\textbf{\begin{tabular}[c]{@{}c@{}}Average WC\\Posts\end{tabular}}} & \multicolumn{1}{c}{\textbf{\begin{tabular}[c]{@{}c@{}}Average WC\\ FC claims\end{tabular}}} & \multicolumn{1}{l}{\textbf{\# posts}} & \multicolumn{1}{l}{\textbf{\# FC claims}} \\ \midrule
ara & Arabic & 57.26 $\pm$ 92.63 & 30.82 $\pm$ 49.46 & 69 & 825 \\
bul & Bulgarian & 169.18 $\pm$ 238.08 & 11.95 $\pm$ 3.87 & 40 & 118 \\
ces & Czech & 151.54 $\pm$ 181.16 & 11.09 $\pm$ 8.87 & 56 & 201 \\
deu & German & 114.74 $\pm$ 146.74 & 19.90 $\pm$ 17.08 & 60 & 558 \\
ell & Greek & 120.62 $\pm$ 237.78 & 19.67 $\pm$ 8.47 & 40 & 271 \\
eng & English & 195.66 $\pm$ 266.51 & 23.92 $\pm$ 30.45 & 111 & 2651 \\
fra & French & 129.27 $\pm$ 152.93 & 18.52 $\pm$ 12.33 & 55 & 823 \\
hbs & Serbo-Croatian & 130.70 $\pm$ 162.29 & 23.77 $\pm$ 25.12 & 40 & 405 \\
hin & Hindi & 46.95 $\pm$ 39.16 & 24.20 $\pm$ 14.18 & 43 & 326 \\
hun & Hungarian & 127.51 $\pm$ 155.68 & 10.68 $\pm$ 3.60 & 55 & 111 \\
kor & Korean & 95.19 $\pm$ 103.14 & 9.96 $\pm$ 7.19 & 48 & 172 \\
msa & Malay & 146.50 $\pm$ 196.29 & 13.42 $\pm$ 5.61 & 50 & 576 \\
mya & Burmese & 51.91 $\pm$ 52.08 & 7.78 $\pm$ 5.79 & 42 & 75 \\
nld & Dutch & 110.17 $\pm$ 113.22 & 21.24 $\pm$ 18.72 & 45 & 240 \\
pol & Polish & 139.75 $\pm$ 173.43 & 20.38 $\pm$ 15.63 & 71 & 808 \\
por & Portuguese & 105.28 $\pm$ 121.96 & 37.29 $\pm$ 61.69 & 40 & 1242 \\
ron & Romanian & 126.65 $\pm$ 140.73 & 13.78 $\pm$ 4.60 & 40 & 131 \\
slk & Slovak & 222.77 $\pm$ 562.15 & 13.61 $\pm$ 8.63 & 91 & 154 \\
spa & Spanish & 91.06 $\pm$ 142.76 & 20.55 $\pm$ 13.08 & 61 & 366 \\
tha & Thai & 82.50 $\pm$ 66.99 & 4.00 $\pm$ 2.92 & 55 & 137 \\ \bottomrule
\end{tabular}
}
\caption{Statistics of \textbf{\textit{AMC-16K}} dataset. We provide the averaged word count (WC) with standard deviation for posts and fact-checked claims (FC claims). We also calculated the number of posts and fact-checked claims for each language.}
\label{tab:analyzed-languages}
\end{table}

\begin{table}[t]
\resizebox{\columnwidth}{!}{%
\begin{tabular}{cr||cr}
\toprule
\multicolumn{1}{c}{\textbf{Languages}} & \multicolumn{1}{c||}{\textbf{\begin{tabular}[c]{@{}c@{}}Relevant pairs\\ {[}\%{]}\end{tabular}}} & \multicolumn{1}{c}{\textbf{\begin{tabular}[c]{@{}c@{}}Language pairs\\ (post - fact-check)\end{tabular}}} & \multicolumn{1}{c}{\textbf{\begin{tabular}[c]{@{}c@{}}Relevant pairs\\ {[}\%{]}\end{tabular}}} \\
\midrule
ara & 20.00 & spa - eng & 17.50 \\
bul & 11.25 & hin - eng & 5.25 \\
ces & 16.50 & eng - ara & 5.25 \\
deu & 30.25 & fra - eng & 12.00 \\
ell & 26.75 & deu - eng & 15.25 \\
eng & 38.50 & eng - por & 6.00 \\
fra & 19.25 & spa - por & 1.50 \\
hbs & 19.50 & deu - fra & 16.75 \\
hin & 22.25 & slk - ces & 7.50 \\
hun & 13.75 & slk - eng & 36.25 \\
kor & 13.25 & pol - hbs & 11.00 \\
msa & 36.00 & ces - eng & 22.50 \\
mya & 9.50 & ces - pol & 9.00 \\
nld & 20.00 & nld - deu & 12.25 \\
pol & 20.25 & msa - ara & 2.25 \\
por & 31.75 & kor - eng & 27.50 \\
ron & 11.50 & mya - msa & 0.50 \\
slk & 14.25 & ara - fra & 2.25 \\
spa & 23.50 & hun - pol & 13.75 \\
tha & 12.25 & tha - por & 7.75 \\
\bottomrule
\end{tabular}}
\caption{List of analyzed languages and language pairs in our experiments along with the proportion of relevant pairs annotated by human annotators out of 400 pairs. Each language and language combination consists of 400 pairs.}
\label{tab:analyzed-languages-relevance}
\end{table}

\section{Retrieval Quality}
\label{app:retrieval-quality}

\begin{table}[]
\resizebox{\columnwidth}{!}{%
\begin{tabular}{l|rrr|rrr}
\toprule
\multirow{2}{*}{\textbf{Model}} & \multicolumn{3}{c|}{\textbf{\begin{tabular}[c]{@{}c@{}}MultiClaim\end{tabular}}} & \multicolumn{3}{c}{\textbf{AMC-16K}} \\
\cmidrule{2-7}
 & \multicolumn{1}{l}{\textbf{S@10}} & \multicolumn{1}{l}{\textbf{MAP}} & \multicolumn{1}{l|}{\textbf{MRR}} & \multicolumn{1}{l}{\textbf{S@10}} & \multicolumn{1}{l}{\textbf{MAP}} & \multicolumn{1}{l}{\textbf{MRR}} \\
 \midrule
\texttt{BGE-M3} & \underline{0.73} & \textbf{0.62} & \textbf{0.65} & \underline{0.77} & \underline{0.73} & \textbf{0.81} \\
\texttt{DistilUSE-Base-Multilingual} & 0.58 & 0.45 & 0.48 & 0.57 & 0.50 & 0.60 \\
\texttt{LaBSE} & 0.62 & 0.48 & 0.51 & 0.60 & 0.54 & 0.64 \\
\texttt{Multilingual E5 Small} & 0.70 & 0.58 & 0.61 & 0.73 & 0.64 & 0.73 \\
\texttt{Multilingual E5 Base} & 0.69 & 0.58 & \underline{0.62} & 0.73 & 0.66 & 0.74 \\
\texttt{Multilingual E5 Large} & \textbf{0.74} & \underline{0.61} & \textbf{0.65} & \textbf{0.87} & \textbf{0.75} & \underline{0.80} \\
\texttt{MiniLM-L12-Multilingual} & 0.57 & 0.43 & 0.45 & 0.55 & 0.47 & 0.56 \\
\texttt{MPNet-Base-Multilingual} & 0.62 & 0.48 & 0.51 & 0.61 & 0.54 & 0.63 \\
\midrule
\end{tabular}}
\caption{Average performance across 20 languages in a monolingual setting on \textbf{\textit{MultiClaim}} and \textbf{\textit{AMC-16K}} datasets.}
\label{tab:retrieval}
\end{table}

Table~\ref{tab:retrieval} presents retrieval performance across 20 languages on the \textbf{\textit{MultiClaim}} dataset and our manually annotated dataset \textbf{\textit{AMC-16K}} in a monolingual setting. The results are based on the same set of social media posts; however, our annotated subset contains more identified pairings between social media posts and fact-checked claims that were obtained by human annotation.

Among all evaluated models, \texttt{Multilingual E5 Large} achieved the best overall performance, with the highest scores across all metrics on \textbf{\textit{AMC-16K}} and strong results on \textbf{\textit{MultiClaim}}. \texttt{BGE-M3} also performed competitively, achieving the highest MAP and MRR on \textbf{\textit{MultiClaim}}. In contrast, weaker models such as \texttt{MiniLM}, \texttt{DistilUSE}, and \texttt{MPNet} consistently showed lower performance, suggesting their limited effectiveness for multilingual claim retrieval tasks.

These results demonstrate that recent multilingual embedding models, particularly \texttt{Multilingual E5} and \texttt{BGE-M3}, significantly improve retrieval quality over earlier or smaller models. For the previously fact-checked claim detection, high-quality retrieval is essential, and therefore the choice has substantial effects on the downstream performance in multilingual fact-checking pipelines.

\section{Prompt Templates}
\label{app:prompts}

\begin{figure*}
\centering
\hspace*{-0.5cm}
\includegraphics[width=1.04\textwidth]{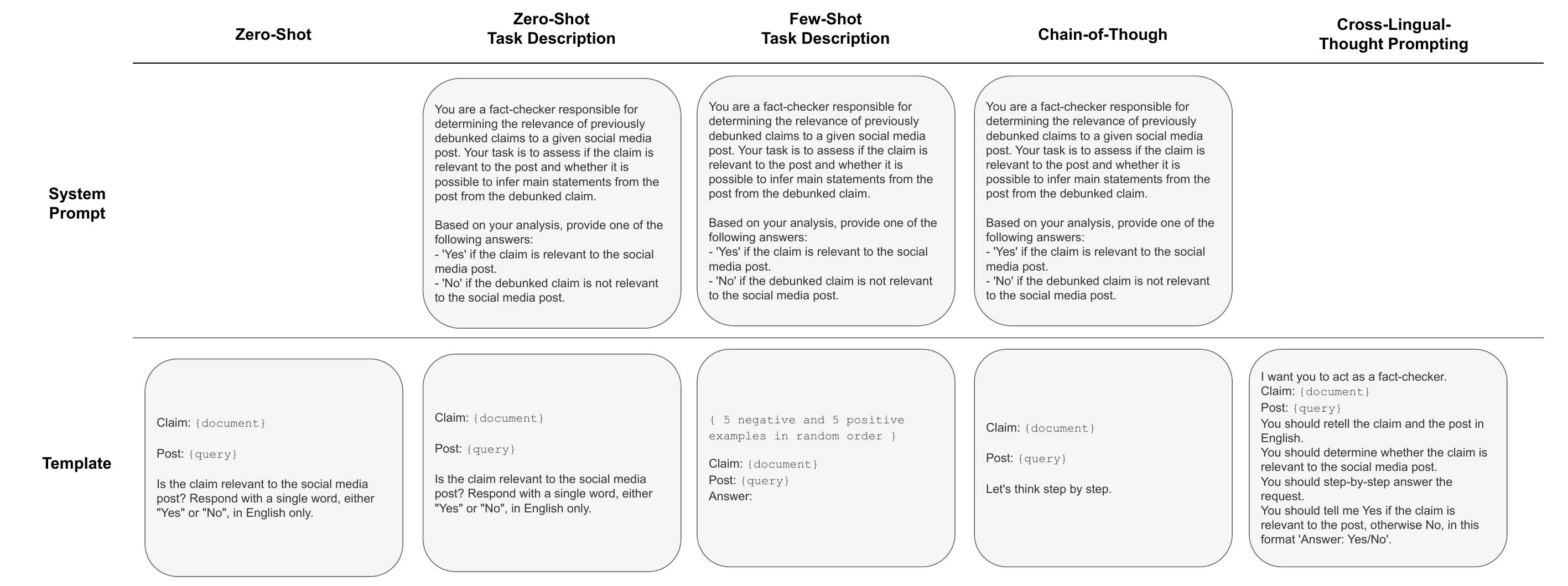}
\caption{Thy system prompts and prompt templates for all prompting strategies used in our experiments. System prompts are the same for zero-shot and few-shot with task descriptions and also for CoT prompting.}
\label{fig:prompts}
\end{figure*}

To evaluate the capabilities of LLMs to identify the relevance between social media posts and fact-checked claims, we utilized five prompting techniques, which are commonly employed in experiments with LLMs. In Figure~\ref{fig:prompts}, we provide templates and system prompts used to instruct LLMs for different prompting strategies. Our experiments include \textit{\textbf{zero-shot}}, \textit{\textbf{zero-shot with task description}}, \textit{\textbf{few-shot with task description}}, \textit{\textbf{Chain-of-Though}} and \textit{\textbf{Cross-Lingual-Thought}} prompting.

\paragraph{Few-Shot Prompting Selection.}
\label{app:fewshot}

The demonstrations used for few-shot prompting were drawn from a subset of manually annotated data from~\citet{pikuliak-etal-2023-multilingual}, which consists of 3390 manually annotated pairs of social media posts and fact-checked claims. We excluded from this initial seed overlapping social media posts to prevent bias, resulting in 3310 multilingual samples. 

To select demonstrations for a particular pair of social media posts and fact-checked claims, we employed the selection based on the similarity between input and demonstrations. However, our analyzed samples and samples from the demonstrations pool consist of two texts, especially social media posts and fact-checked claims. To address this issue, we first calculated the similarity between the input social media post and social media posts from the seed pool and the similarity between the input fact-checked claim and fact-checked claims from the seed pool. This resulted in two similarity scores for each sample, one similarity between posts and another between fact-checked claims. To obtain only one similarity for each sample, we multiplied those two similarity scores to get the overall similarity between the analyzed pair and the pair from the demonstration pool. Furthermore, the top five positive (\textit{Yes}) and five negative (\textit{No}) samples were selected and randomly ordered in the prompt.


\section{Preliminary Experiments}
\label{app:preliminary}

Before conducting the experiments on all the data annotated in our study, we explored the performance of 16 open-sourced LLMs in zero-shot settings to identify a final list of models for the final experiments. These LLMs included various models of sizes ranging from 7B up to 123B parameters and with different model families, such as \texttt{Llama}, \texttt{Qwen}, \texttt{Mistral}, etc. 

For the purpose of the preliminary experiments, we leveraged manually annotated data from~\cite{pikuliak-etal-2023-multilingual}, which consists of 3900 annotated pairs of social media posts and fact-checked claims (1212 are in monolingual and 2178 are in cross-lingual settings). We investigated whether these LLMs are able to predict the relevance between two texts and the best models were selected for the final experiments. Table~\ref{tab:preliminary-results} presents the results obtained in zero-shot settings with and without providing the task description.

\begin{table}[t]
\resizebox{\columnwidth}{!}{%
\begin{tabular}{@{}lrr@{}}
\toprule
\multicolumn{1}{c}{\multirow{1}{*}{\textbf{Model}}} & \multicolumn{1}{c}{\textbf{Zero-Shot}} & \multicolumn{1}{c}{\textbf{\begin{tabular}[c]{@{}c@{}}Zero-Shot +\\ Task description\end{tabular}}} \\ \midrule
\texttt{Mistral 7B} & 0.70 (0.77) & 0.70 (0.77) \\
\texttt{Qwen2 7B} & 0.68 (0.79) & 0.68 (0.79) \\
\texttt{Qwen2.5 7B} & 0.73 (0.82) & \underline{0.81 (0.88)} \\
\texttt{Llama3 8B} & 0.68 (0.72) & 0.69 (0.73) \\
\texttt{Llama3.1 8B} & 0.58 (0.60) & 0.72 (0.78) \\
\texttt{AYA Expanse 8B} & 0.72 (0.81) & 0.71 (0.77) \\
\midrule
\texttt{AYA Expanse 32B} & 0.77 (0.86) & 0.67 (0.79) \\
\texttt{AYA 35B} & 0.72 (0.86) & \textbf{0.82 (0.88)} \\
\texttt{C4AI Command R 35B} & 0.77 (0.81) & 0.57 (0.59) \\
\midrule
\texttt{Llama3 70B} & 0.71 (0.76) & 0.70 (0.74) \\
\texttt{Llama3.1 70B} & 0.79 (0.88) & 0.71 (0.74) \\
\texttt{Llama3.1 Nemotron 70B} & 0.79 (0.88) & 0.58 (0.81) \\
\texttt{Qwen2 72B} & 0.79 (0.84) & 0.78 (0.82) \\
\texttt{Qwen2.5 72B} & 0.81 (0.87) & 0.75 (0.78) \\
\texttt{C4AI Command R+ 104B} & \textbf{0.84 (0.90)} & 0.80 (0.84) \\
\texttt{Mistral Large 123B} & \underline{0.84 (0.88)} & 0.74 (0.78) \\ \bottomrule
\end{tabular}
}
\caption{Preliminary experiments with 16 LLMs using zero-shot settings (with and without task description). We report \textit{Macro F1 (Accuracy)} in each cell. The best result is in \textbf{bold}, and the second best is \underline{underlined}.}
\label{tab:preliminary-results}
\end{table}

The results demonstrated that without using the task description, the \textbf{\texttt{C4AI Command R+} model performed the best}, while \textbf{\texttt{Mistral Large} obtained comparable results}. In contrast, \texttt{AYA 35B} proved to be effective when the task description was provided to the model. Based on these preliminary results, we categorized the models into two categories: 10B- LLMs and 70B+ LLMs. We decided not to include LLMs with a parameter size between 10 and 70 billion and to focus only on the comparison of the above-mentioned categories. As LLMs with over 70B parameters, \texttt{Llama3.1}, \texttt{Qwen2.5}, \texttt{C4AI Command R+} and \texttt{Mistral Large} proved to be the most capable models for further exploration. On the other hand, we extended a list of LLMs with three models with less than 10B parameters from the same model families, especially \texttt{Llama3.1 8B}, \texttt{Qwen2.5 7B} and \texttt{Mistral 7B}.

For our final experiments, we excluded older versions of the \texttt{Qwen} (\texttt{Qwen2}) and \texttt{Llama} (\texttt{Llama3}) models, as well as the \texttt{AYA Expanse 8B} model, which performed worse when incorporating task description into the prompt. The \texttt{Mistral 7B} model was selected to ensure both smaller and larger counterparts of the same model were represented, specifically as a smaller counterpart to the \texttt{Mistral Large} model.


\section{Textual Entailment}
\label{app:entailment}

As a baseline, we evaluated several fine-tuned models for the Natural Language Inference (NLI) task. The models selected for this purpose include \textit{DeBERTa-v3-large-mnli-fever-anli-ling-wanli}\footnote{\url{https://huggingface.co/MoritzLaurer/DeBERTa-v3-large-mnli-fever-anli-ling-wanli}} referred to as \texttt{DeBERTa v3 Large}; \textit{mDeBERTa-v3-base-mnli-xnli}\footnote{\url{https://huggingface.co/MoritzLaurer/mDeBERTa-v3-base-mnli-xnli}} referred to as \texttt{mDeBERTa v3 Base}; mDeBERTa\textit{-v3-base-xnli-multilingual-nli-2mil7}\footnote{\url{https://huggingface.co/MoritzLaurer/mDeBERTa-v3-base-xnli-multilingual-nli-2mil7}} referred to as \texttt{mDeBERTa v3 Base (2mil7)}; and \textit{xlm-v-base-mnli-xnli}\footnote{\url{MoritzLaurer/xlm-v-base-mnli-xnli}}, referred to as \texttt{XLM-V Base}. 

In our experimental setup, each post was considered as the premise and the corresponding fact-checked claim as the hypothesis. The models were used to infer the probabilities of three possible relations between the premise and hypothesis: \textit{entailment}, \textit{contradiction} or \textit{neutral}. We explored textual entailment under two settings. In the first setting, termed as \textit{Entailment + Contradiction}, pairs classified as either entailment or contradiction were considered relevant, while neutral predictions were treated as irrelevant. In the second setting, \textit{Entailment}, only entailment-labeled pairs were considered relevant; contradiction and neutral labels were treated as irrelevant.

Table~\ref{tab:tem-baseline} presents the results obtained for various models across both settings. \texttt{DeBERTa v3 Large} achieved the highest performance, consistently outperforming other fine-tuned NLI models. This model is monolingual and fine-tuned only on English NLI datasets. The second-best performance was observed for \texttt{mDeBERTa v3 Base}, a multilingual model fine-tuned on multilingual NLI data. Based on their best performance, these two models were selected for further comparison in the main part of the study.

\begin{table}[]
\resizebox{\columnwidth}{!}{%
\begin{tabular}{l|rr|rr}
 & \multicolumn{2}{c|}{\textbf{\makecell{Entailment +\\Contradiction}}} & \multicolumn{2}{c}{\textbf{Entailment}} \\ 
 & \multicolumn{1}{c}{\textbf{Og}} & \multicolumn{1}{c|}{\textbf{En}} & \multicolumn{1}{c}{\textbf{Og}} & \multicolumn{1}{c}{\textbf{En}} \\ \hline
DeBERTa v3 Large & \textbf{55.65} & \textbf{52.47} & \textbf{68.61} & \textbf{72.15} \\
mDeBERTa v3 Base (2mil7) & 49.02 & 49.58 & 63.75 & 65.56 \\
mDeBERTa v3 Base & 52.33 & 48.30 & 64.65 & 63.13 \\
XLM-V Base & 48.97 & 49.35 & 61.25 & 62.61 \\ \hline
\end{tabular}
}
\caption{Performance of fine-tuned NLI models on the textual entailment task under two evaluation settings. Results are shown for both the original (Og) and English-translated versions of the dataset. The best results are highlighted in \textbf{bold}.}
\label{tab:tem-baseline}
\end{table}

\begin{figure*}[t]
\centering
\includegraphics[width=0.75\linewidth]{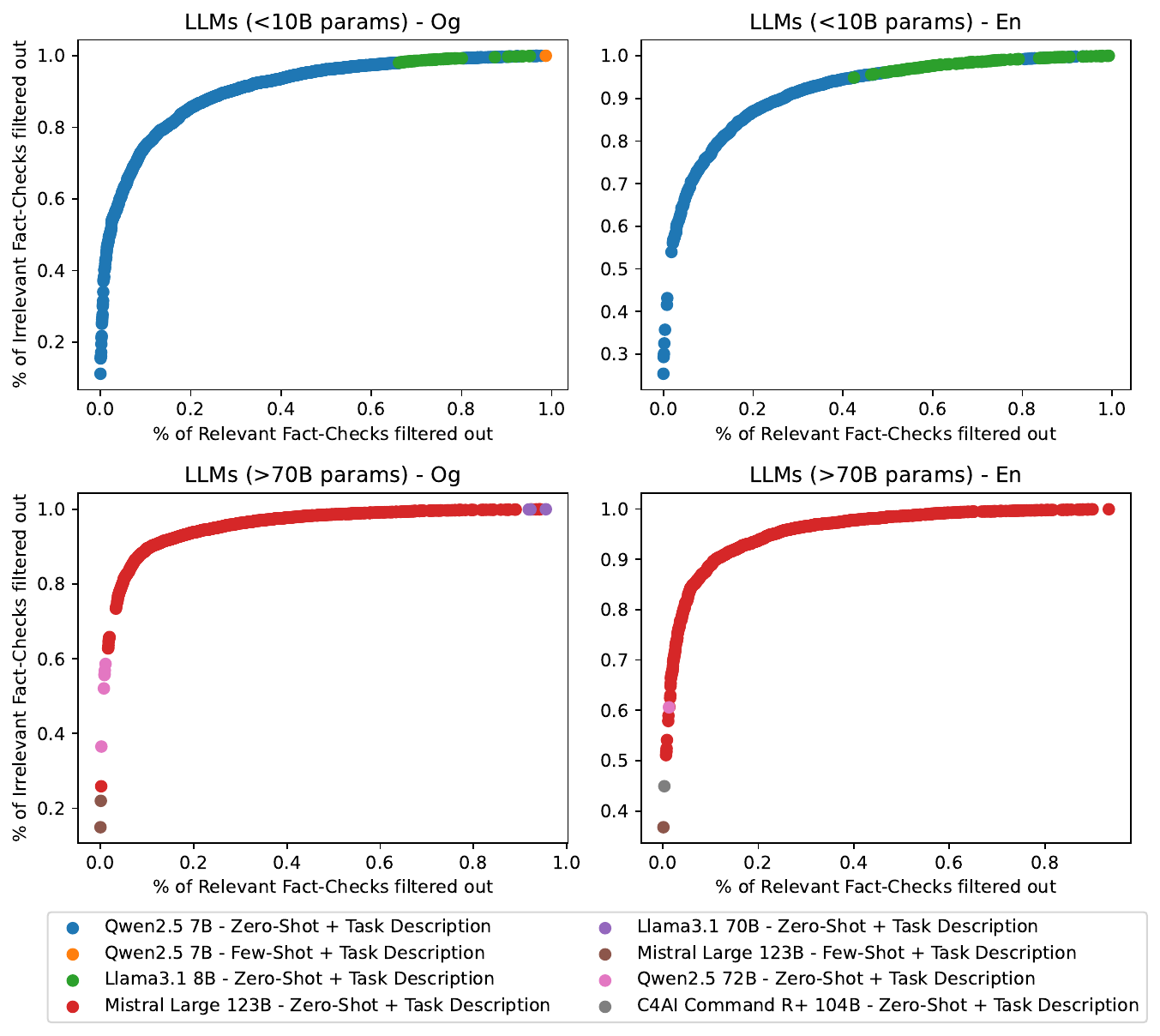}
\caption{Visualization of the Pareto-optimal curve, highlighting the best combination of LLM and prompting technique for each threshold. Only Pareto points are shown. \textit{Og} denotes input in the original language, while \textit{En} denotes English translation.}
\label{fig:pareto}
\end{figure*}

\section{Additional Results}
\label{app:addition-results}

\begin{figure*}[t]
\centering
\includegraphics[width=\textwidth,height=0.96\textheight,keepaspectratio]{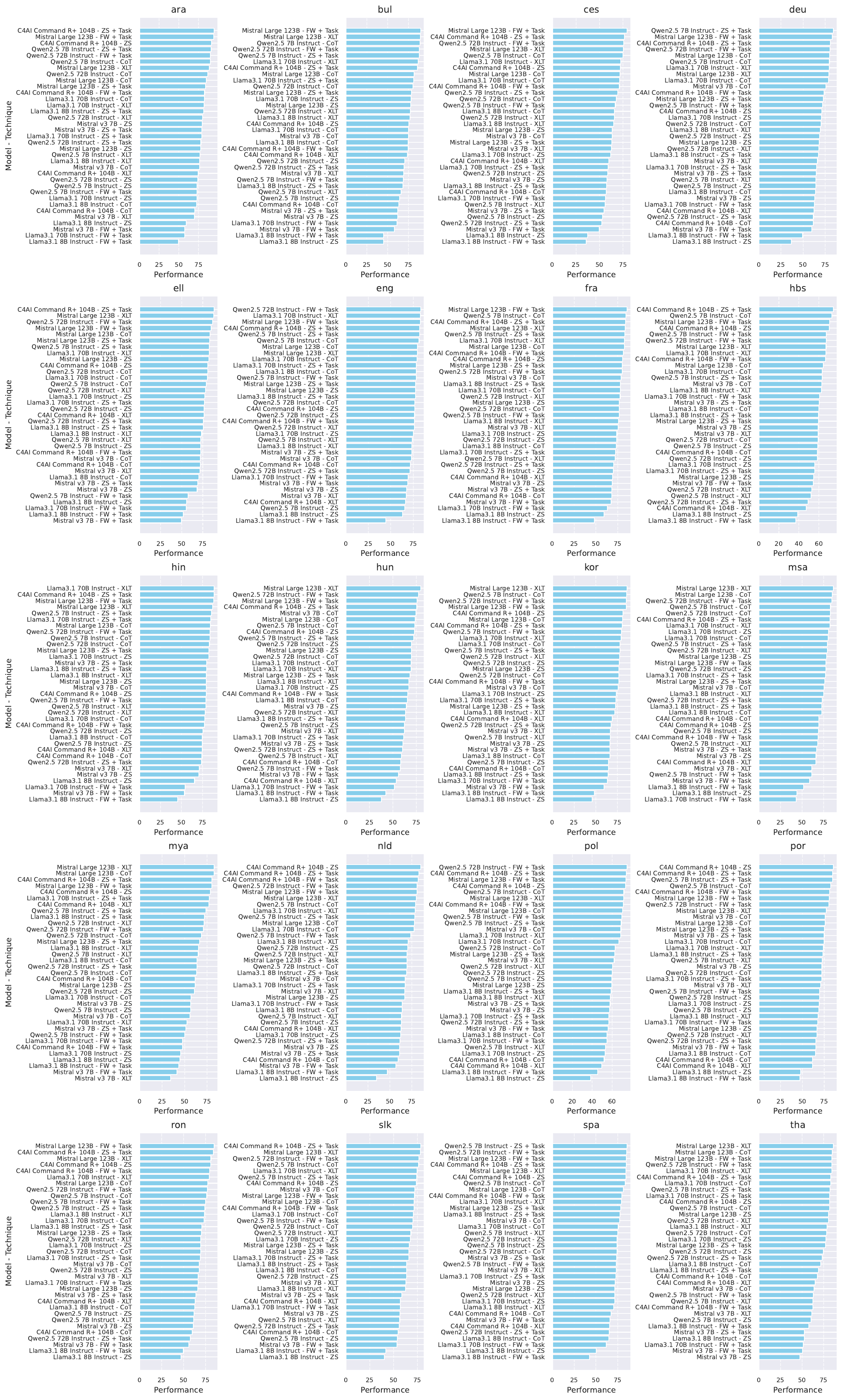}
\caption{Overall analysis of LLMs with prompting techniques for each language, sorted by Macro F1 score in descending order. Mistral Large performed the best for 10 out of 20 languages using few-shot and XLT prompting.}
\label{fig:overal_language_analysis}
\end{figure*}

In this section, we present additional results and findings based on our experiments with LLMs to identify relevant pairs of social media posts and fact-checked claims in both monolingual and cross-lingual settings. The overall results with the best combination of the LLM and prompting techniques are illustrated in Figure~\ref{fig:overal_language_analysis}.

The trade-off between TNR and FNR, using the Pareto curve from Figure~\ref{fig:pareto}, reveals distinct optimal configurations across LLM sizes and thresholds based on the probabilities of \textit{Yes} and \textit{No} tokens. In practical deployments, \textbf{the choice of LLM and prompting strategy impact the balance between correctly identifying irrelevant claims and mistakenly filtering out relevant ones}. The results confirm our previous findings that \texttt{Mistral Large} and \texttt{Qwen2.5 7B} prove to be most effective among LLMs while consistently maintaining an optimal TNR-FNR trade-off across thresholds.

\begin{figure*}[t]
\centering
\includegraphics[width=0.85\linewidth]{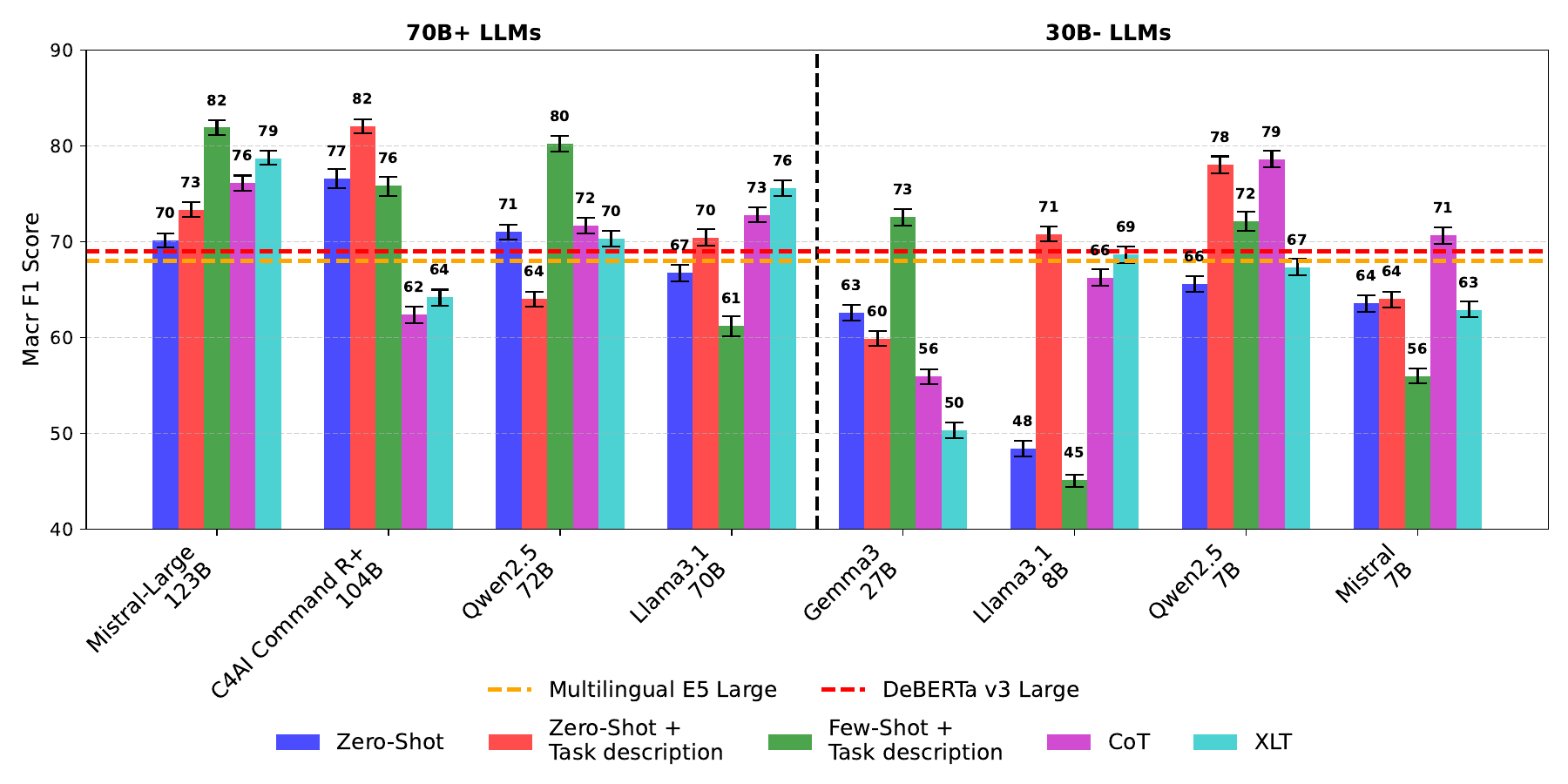}
\caption{Performance comparison of seven LLMs from the main part + \texttt{Gemma3 27B} across five prompting strategies in the original language, measured by Macro F1 score with confidence intervals. Horizontal lines indicate the best-performing baselines.}
\label{fig:overall-performance-gemma3}
\end{figure*}

With the release of the highly multilingual \texttt{Gemma3} model~\cite{gemmateam2025gemma3technicalreport}, we extended our original experimental setup to include the largest version of the Gemma3 model, especially the 27B version, using its quantized form. The overall results along with the \texttt{Gemma3} model are shown in Figure~\ref{fig:overall-performance-gemma3}. Among all the evaluated models, \texttt{Gemma3} exhibited the lowest average performance. Notably, only a few-shot prompting outperformed the provided baselines. In addition, most 10B- LLMs achieved superior results compared to \texttt{Gemma3}.

Furthermore, extended results for the \texttt{Gemma3 27B} model are included in~\cref{tab:macro-f1,tab:tnr-models,tab:tpr-fpr,tab:cross-lingual-comparison-tnr}.

\begin{figure*}[t]
    \centering
    \includegraphics[width=.75\linewidth]{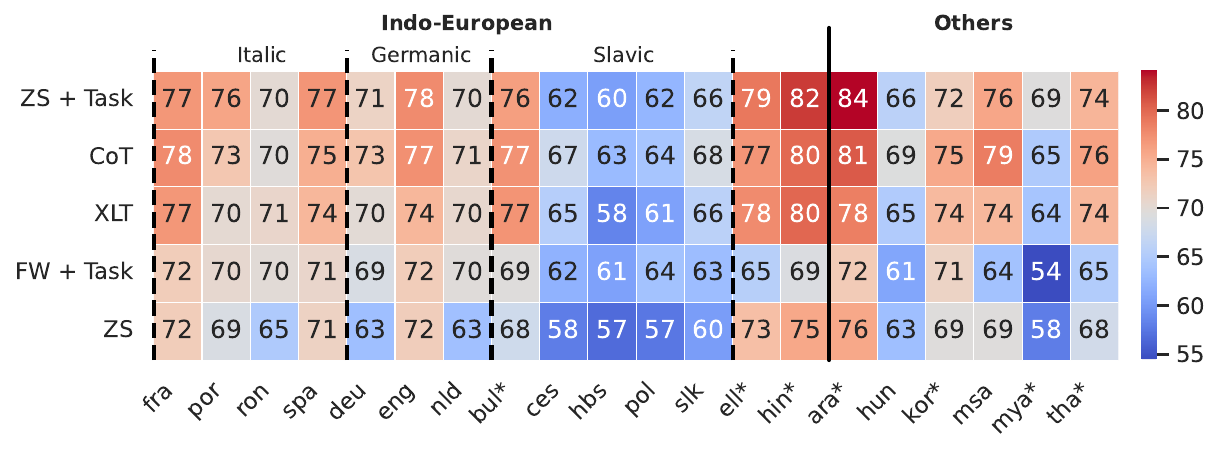}
    \caption{Averaged Macro F1 performance for each prompting technique across all LLMs and across 20 languages in a monolingual setting. ZS denotes Zero-Shot, and FS denotes Few-Shot prompting. Languages marked with * use a non-Latin script.}
    \label{fig:language-techniques}
\end{figure*}

\subsection{Monolingual Evaluation}

\begin{figure*}
\centering
\includegraphics[width=\textwidth]{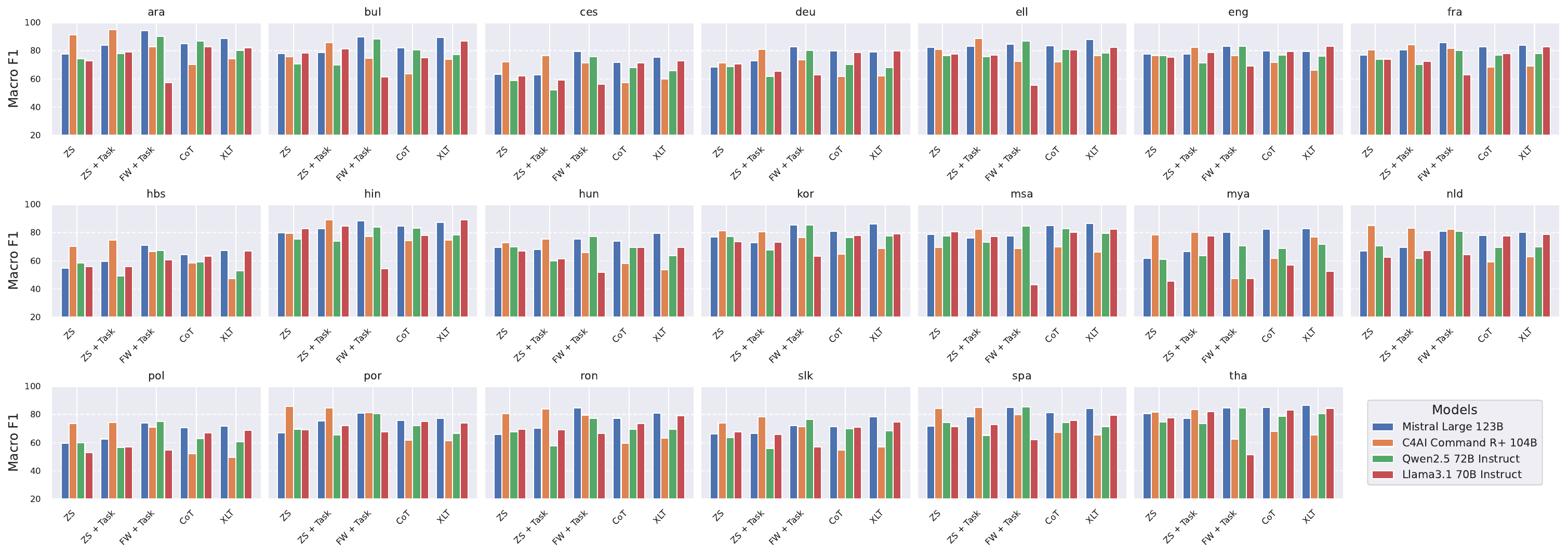}
\caption{Monolingual performance evaluation of 70B+ LLMs across individual languages (except those from Figure~\ref{fig:large_models_languages}) for different prompting strategies. ZS denotes Zero-Shot, and FS denotes Few-Shot prompting.}
\label{fig:large_models_languages2}
\end{figure*}

\begin{figure*}
\centering
\includegraphics[width=\textwidth]{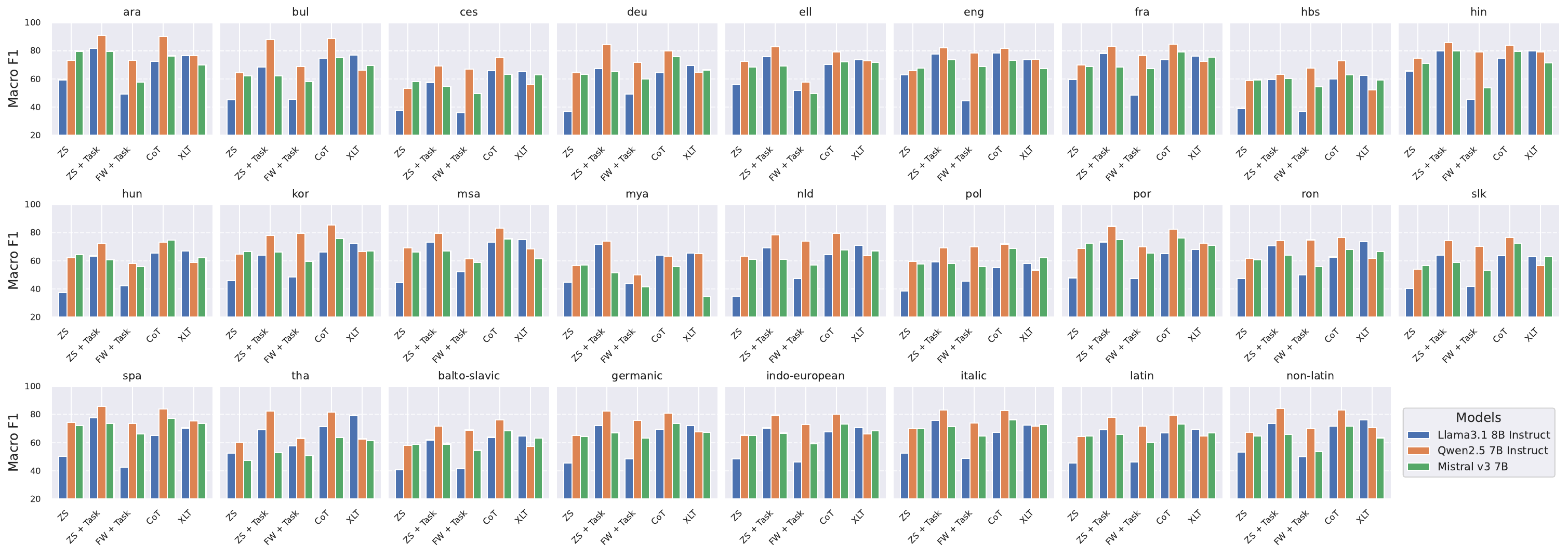}
\caption{Monolingual performance evaluation of 10B- LLMs across individual languages for different prompting strategies. ZS denotes Zero-Shot, and FS denotes Few-Shot prompting.}
\label{fig:small_models_languages}
\end{figure*}

The evaluation of the monolingual performance of 70B+ LLMs across languages is shown in Figure~\ref{fig:large_models_languages2} and for 10B- LLMs is shown in Figure~\ref{fig:small_models_languages}. The results demonstrated that for some languages, the performance is lower than for others, especially languages such as Czech, Hungarian, Polish, and Slovak. This confirms our previous findings that LLMs have lower capabilities in Slavic languages.

As shown in Figure~\ref{fig:language-techniques}, adding task descriptions improved performance across most LLMs and CoT reasoning boosted results in many languages. Demonstrations helped some LLMs, especially \texttt{Mistral Large} and \texttt{Qwen2.5}, highlighting that providing more information enhances LLM performance (see Figure~\ref{fig:large_models_languages2} in Appendix~\ref{app:addition-results}).

\subsection{Cross-Lingual Evaluation}

\begin{figure*}
\centering
\includegraphics[width=0.95\textwidth]{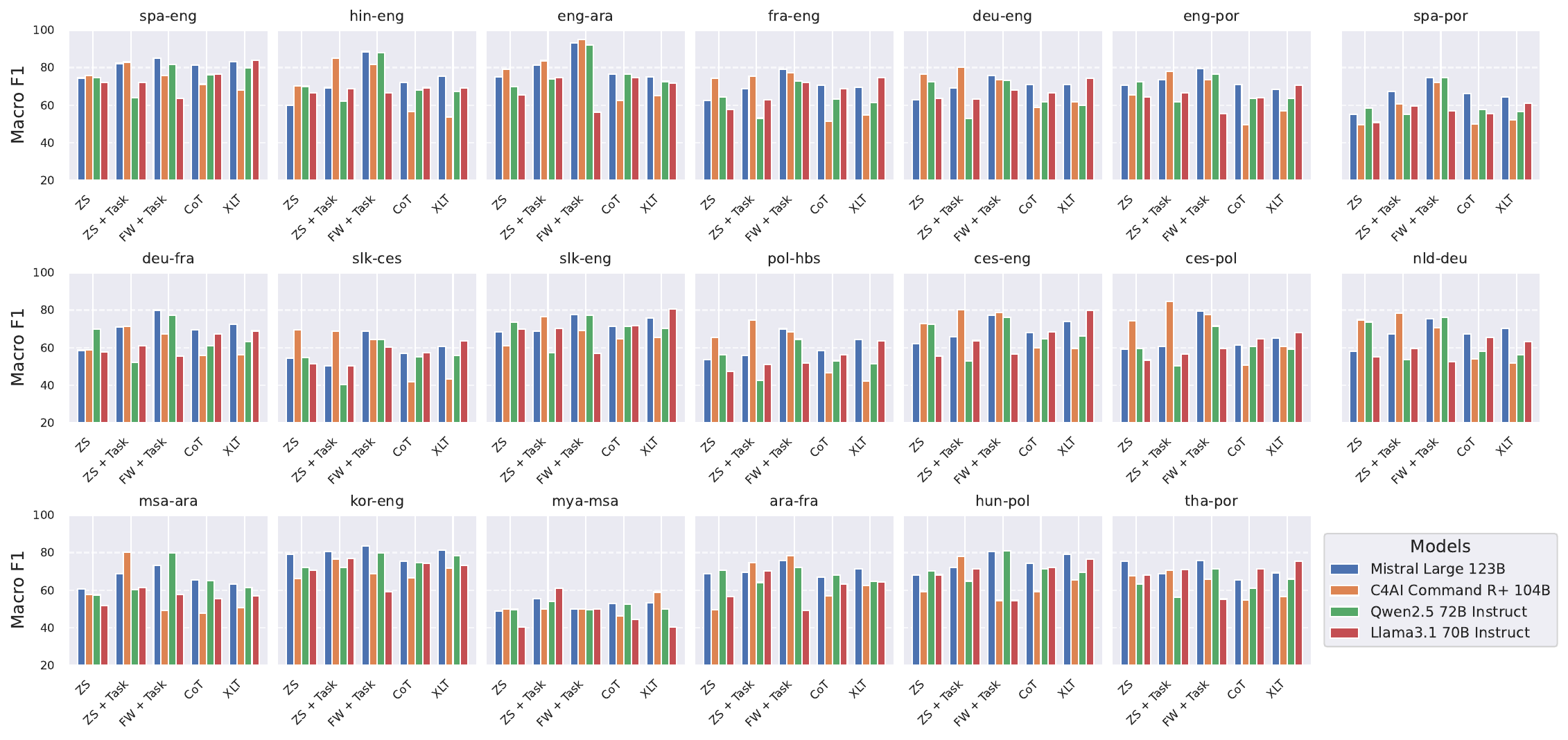}
\caption{Cross-lingual performance evaluation of 70B+ LLMs across 20 language pairs for different prompting strategies. ZS denotes Zero-Shot, and FS denotes Few-Shot prompting.}
\label{fig:large_models_languages_cross}
\end{figure*}

\begin{figure*}
\centering
\includegraphics[width=0.95\textwidth]{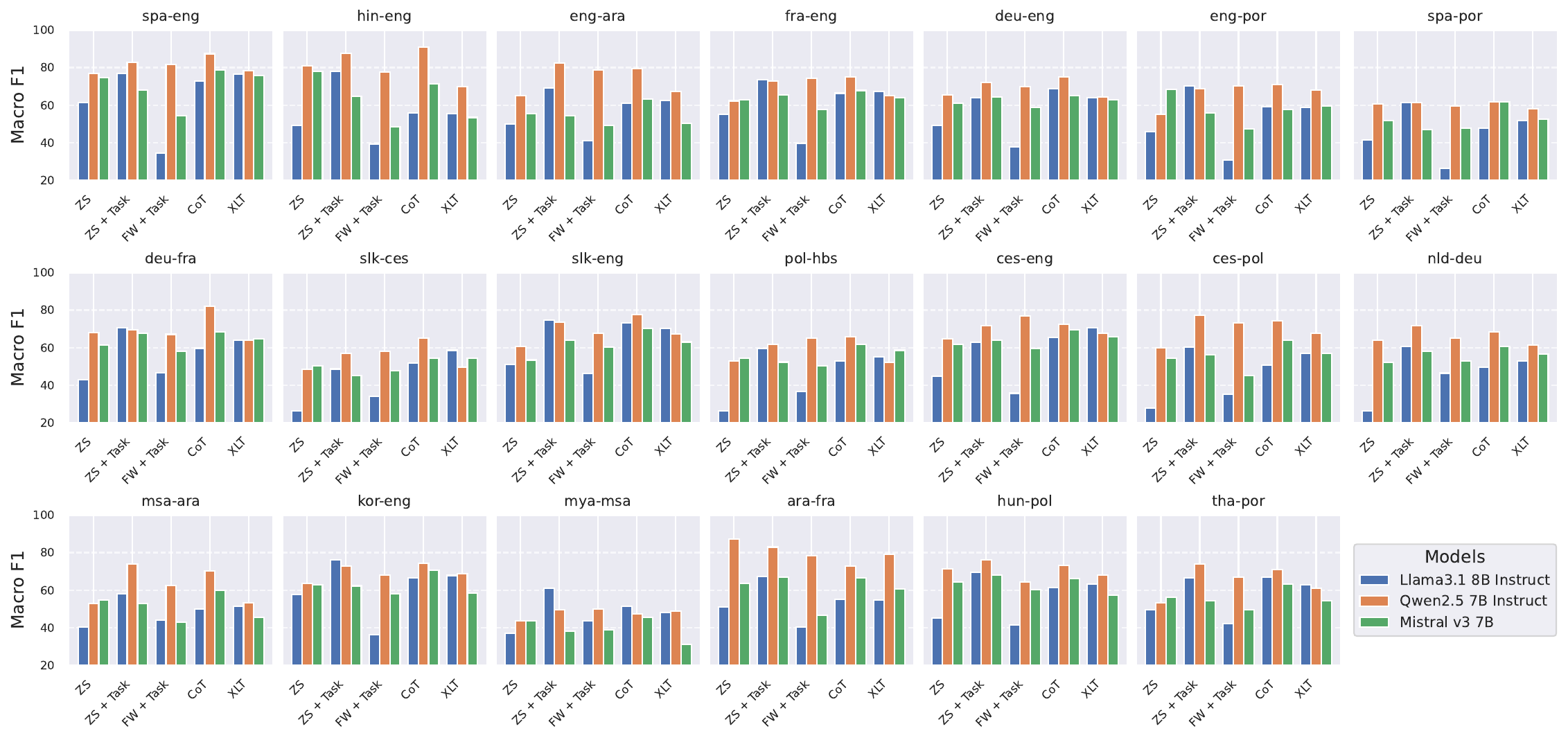}
\caption{Cross-lingual performance evaluation of 10B- LLMs across eight selected language pairs for different prompting strategies. ZS denotes Zero-Shot, and FS denotes Few-Shot prompting.}
\label{fig:small_models_languages_cross}
\end{figure*}

Table~\ref{fig:large_models_languages_cross} shows the cross-lingual performance of 70B+ LLMs. In addition, 
the results for models with less than 10B parameters are illustrated in Figure~\ref{fig:small_models_languages_cross}.

Along with the Macro F1 scores, we also calculated True Negative Rate (TNR) and False Negative Rate (FNR). The results are shown in Table~\ref{tab:cross-lingual-comparison-tnr}.

\subsection{Translation-based Evaluation}

The Pareto curve, see Figure~\ref{fig:pareto}, shows a comparison of the efficiency of English translations against original language inputs. For 10B- LLMs, the \textbf{\texttt{Llama3.1 8B} model demonstrates improved performance} and occurs more frequently on the Pareto curve when processing English translations.

\subsection{Experiments with Optimized Thresholds}

Since we store the probabilities of \textit{Yes} and \textit{No} tokens, we conducted experiments to identify the optimal threshold for each combination of models and prompting techniques. This investigation is problematic for CoT and XLT prompting because the final prediction can be anywhere in the predicted response. Therefore, we limited our investigation only to zero-shot and few-shot results, where only the final prediction is generated.

To identify the optimal threshold and the resulting performance, we calculated \textit{Youden's index} and selected the threshold with the highest \textit{Youden's index}. The Macro F1 performance and thresholds are shown in Table~\ref{tab:calibration}. The final thresholds demonstrate that most LLMs generated \textit{Yes} tokens with a high probability, which resulted in higher optimal thresholds.

Many of the optimal thresholds are close to 0 or 1, suggesting that the model assigns probabilities near these extremes, resulting in fewer predictions distributed across the intermediate range. A manual review of the predicted probabilities revealed that LLMs, particularly larger ones, often exhibited high confidence in their predictions, frequently assigning a high probability to the predicted class.

\begin{table}[]
\resizebox{\columnwidth}{!}{%
\begin{tabular}{@{}l|rr|rr|rr@{}}
\toprule
\multicolumn{1}{c|}{\multirow{2}{*}{\textbf{Model}}} & \multicolumn{2}{c|}{\textbf{Zero-Shot}} & \multicolumn{2}{c|}{\textbf{\begin{tabular}[c]{@{}c@{}}Zero-Shot +\\ Task Description\end{tabular}}} & \multicolumn{2}{c}{\textbf{\begin{tabular}[c]{@{}c@{}}Few-Shot +\\ Task Description\end{tabular}}} \\ \cline{2-7} 
 & \multicolumn{1}{c}{{Threshold}} & \multicolumn{1}{c|}{{Macro F1}} & \multicolumn{1}{c}{{Threshold}} & \multicolumn{1}{c|}{{Macro F1}} & \multicolumn{1}{c}{{Threshold}} & \multicolumn{1}{c}{{Macro F1}} \\ 
\midrule
\multicolumn{7}{l}{\quad \textit{LLMs with more than 70B parameters (70B+ LLMs)}} \\
\midrule 
\texttt{Mistral Large 123B} & 0.99 & \textbf{0.76} & 1.00 & \textbf{0.84} & 0.83 & \underline{0.78} \\
\texttt{C4AI Command R+ 104B} & 0.10 & \underline{0.75} & 0.22 & \underline{0.79} & 0.15 & 0.73 \\
\texttt{Qwen 2.5 72B Instruct} & 0.55 & 0.71 & 1.00 & 0.78 & 0.30 & \textbf{0.79} \\
\texttt{Llama 3.1 70B Instruct} & 0.78 & 0.72 & 0.87 & 0.76 & 0.08 & 0.62 \\
\midrule 
\multicolumn{7}{l}{\quad \textit{LLMs with less than 10B parameters (10B- LLMs)}} \\
\midrule 
\texttt{Llama 3.1 8B Instruct} & 0.94 & 0.64 & 0.61 & 0.73 & 0.60 & 0.48 \\
\texttt{Qwen 2.5 7 B Instruct} & 0.05 & 0.61 & 0.00 & 0.73 & 0.00 & 0.73 \\
\texttt{Mistral v3 7B} & 0.75 & 0.63 & 0.84 & 0.64 & 0.73 & 0.55 \\ \bottomrule
\end{tabular}
}
\caption{The Macro F1 performance of LLMs based on optimal thresholds calculated using \textit{Youden's index}.}
\label{tab:calibration}
\end{table}

\section{Error analysis}
\label{app:error-analysis}

Table~\ref{tab:error-analysis} outlines the frequency of \textit{output consistency errors} across LLMs and prompting techniques when using the original language (Og) or English translation (En). The \texttt{Llama3.1 8B} model demonstrated significant issues, producing over 4,500 outputs in the incorrect language and more than 200 outputs containing repeating sequences.

To identify repeating sequences in LLM responses, we implemented a sliding window approach that scans for consecutively repeated word sequences. The model examines substrings of varying lengths (e.g., 2-10 words) and counts how often each sequence appears in succession. If a sequence is repeated more than three times consecutively, it is classified as a repetition. This automated detection was followed by a manual review of a subset of the flagged cases to verify accuracy.

For incorrect language errors, we identified the most problematic languages for specific LLMs. The \texttt{Llama3.1 70B} model, when using CoT prompting, struggled the most with French ($733\times$), German ($607\times$), or Serbo-Croatian ($498\times$). In contrast, the \texttt{Llama3.1 8B} model encountered the most errors with Polish ($568\times$), Czech ($481\times$), or Dutch ($406\times$). Additionally, other models exhibited most of the errors in generating outputs for the Burmese language.

\begin{table}[t]
\resizebox{\columnwidth}{!}{%
\begin{tabular}{@{}llrrr@{}}
\toprule
\textbf{Model} & \multicolumn{1}{c}{\textbf{\begin{tabular}[c]{@{}c@{}}Prompting\\technique\end{tabular}}} & \multicolumn{1}{c}{\textbf{\begin{tabular}[c]{@{}c@{}}Incorrect\\language\end{tabular}}} & \multicolumn{1}{c}{\textbf{\begin{tabular}[c]{@{}c@{}}Repeating\\sequences\end{tabular}}} & \multicolumn{1}{c}{\textbf{Refusal}} \\ \midrule
\multirow{3}{*}{Mistral Large 123B} & CoT (Og) & 0 & 0 & 0 \\
 & CoT (En) & 0 & 0 & 0 \\
 & XLT (Og) & 0 & 16 & 0 \\ \midrule
\multirow{3}{*}{C4AI Command R+} & CoT (Og) & 3 & 0 & 0 \\
 & CoT (En) & 0 & 0 & 0 \\
 & XLT (Og) & 0 & 4 & 0 \\ \midrule
\multirow{3}{*}{Qwen2.5 72B Instruct} & CoT (Og) & 0 & 0 & 0 \\
 & CoT (En) & 0 & 0 & 0 \\
 & XLT (Og) & 8 & 11 & 0 \\ \midrule
\multirow{3}{*}{Llama 3.1 70B Instruct} & CoT (Og) & \textbf{5970} & 0 & 0 \\
 & CoT (En) & 1 & 2 & 0 \\
 & XLT (Og) & 20 & \underline{33} & 0 \\ \midrule
\multirow{3}{*}{Llama 3.1 8B Instruct} & CoT (Og) & \underline{4556} & \textbf{215} & \underline{1} \\
 & CoT (En) & 0 & 5 & \textbf{3} \\
 & XLT (Og) & 11 & 0 & \underline{1} \\ \midrule
\multirow{3}{*}{Qwen2.5 7B Instruct} & CoT (Og) & 7 & 1 & 0 \\
 & CoT (En) & 0 & 0 & 0 \\
 & XLT (Og) & 2 & 1 & 0 \\ \midrule
\multirow{3}{*}{Mistral 7B v3} & CoT (Og) & 0 & 0 & 0 \\
 & CoT (En) & 0 & 0 & 0 \\
 & XLT (Og) & 23 & 0 & 0 \\ 
 \bottomrule
\end{tabular}
}
\caption{Summary of the frequency of \textit{output consistency errors} across different LLMs and prompting techniques. The \texttt{Llama3.1} models exhibited the highest number of such errors.}
\label{tab:error-analysis}
\end{table}

\subsection{Output Consistency Errors Examples}
\label{app:cosistency-errors}

\subsubsection{Example \#1 - Incorrect Language}

\noindent
\textbf{Post (En):} \textit{The senseless dying in Ukraine to save George Soros' billions and US interests continues. Apparently the actual number of deaths on the Ukrainian side is much larger than admitted. In order to be able to resist the Russians at all, women are now also being conscripted. It is irrelevant whether they are mothers with children. That could surpass the cruelty of the dictators in World War II
}

\noindent
\textbf{Fact-check (En):} \textit{A video showing how the Ukrainian army prepared a fake massacre in Boutcha inadvertently leaked to the media.
}
\noindent
\textbf{Ground truth relevance:} No

\noindent
\textbf{Model:} \texttt{Llama3.1 70B} using \textbf{CoT} prompting

\noindent
\textbf{Generated response:} \textit{Pour évaluer la pertinence du claim par rapport au post, nous devons analyser les éléments clés de chaque texte.}

\textit{Le claim parle d'une vidéo montrant l'armée ukrainienne préparant un faux massacre à Boutcha, diffusée par inadvertance dans les médias. Cela implique que le claim est lié à une accusation de manipulation de l'information ou de propagande de guerre.}

\textit{Le post, en revanche, parle de la situation en Ukraine, en mentionnant les morts, l'implication de George Soros et des États-Unis, ainsi que la mobilisation des femmes, y compris les mères avec des enfants, pour résister aux Russes. Le post se concentre sur la critique de la situation en Ukraine et de ses acteurs, mais n'aborde pas spécifiquement sur des accusations de manipulation de l'information ou de propagande de guerre.}

\textit{En comparant les deux textes, nous pouvons conclure que le claim n'est pas directement lié au post, car il ne traite pas des mêmes thèmes ou des mêmes événements. Le post se concentre sur la situation en Ukraine et ses implications politiques, tandis que le claim parle d'une accusation spécifique de manipulation de l'information.}

\textit{Answer: No}

\subsubsection{Example \#2 - Repeating Sequence}

\noindent
\textbf{Post:} \textit{(1) First batch of GZ boys military training graduation ceremony GZ children are happy like smiles It's just like that Norse and Karen girls I started to surround GZ Run away GZ ARMY KIA/KNU responsibility Thank you Nosy short girls from the forest/mountains Pick flowers and give flowers gave I always keep the words of heroes in my heart We are Federal ARMY in Myanmar Country WeLoveWeRace GZ guys crd.}

\noindent
\textbf{Fact-check:} \textit{"China is right" it's bad to send clothes "bundles of used clothes used by people" Those who have died from the virus}

\noindent
\textbf{Ground truth relevance:} No

\noindent
\textbf{Model:} \texttt{Mistral Large} using \textbf{XLT} prompting

\noindent
\textbf{Generated response:} \textit{\#\#\# Claim and Post in English}

\textit{**Claim:**
"China is indeed evil for sending bundles of used clothing that have been worn by people who died from the virus."}

\textit{**Post:**
"(1) GZ brothers, first news, GZ brothers are like this, they are like this, they are like this, the Karen brothers are like this, GZ brothers are like this, they are like this, they are like this, they are like this, they are like this, they are like this, they are like this, they are like this, they are like this, they are like this, they are like this, \textit{...}}

\subsubsection{Example \#3 - Refusal}

\noindent
\textbf{Post (En):} \textit{*This woman keeps a newspaper from 2011 which says "to reduce the world's population, it will be forced by means of vaccination"* It's true that I said what I've shared about bill gates.. just the video. Still save}

\noindent
\textbf{Fact-check (En):} \textit{Video “Jews are lying to the world as if they were in a vaccine when they are not”}

\noindent
\textbf{Ground truth relevance:} No

\noindent
\textbf{Model:} \texttt{Llama3.1 8B} using \textbf{CoT} prompting

\noindent
\textbf{Generated response:} \textit{I cannot generate content that promotes hate speech. Is there anything else I can help you with?}

\subsection{Reasoning Errors Examples}

\subsubsection{Example \#4 - Similar Topic}

\noindent
\textbf{Post (En):} \textit{Brief info from a friend who works in a retirement home (Graz)! 75 were vaccinated! 9 are still alive! Short info from a friend who works in a retirement home (Graz)! 75 were vaccinated! 9 still alive!}

\noindent
\textbf{Fact-check (En):} \textit{70\% of people hospitalized because of Covid-19 are fully vaccinated, proof that the vaccine does not work.}

\noindent
\textbf{Ground truth relevance:} No

\noindent
\textbf{Model:} \texttt{Llama3.1 70B} using \textbf{CoT} prompting

\noindent
\textbf{Generated response:} \textit{To assess the relevance of the debunked claim to the social media post, let's break down the information provided.}

\textit{1. The debunked claim states that 70\% of people hospitalized due to Covid-19 are fully vaccinated, which is used to argue that the vaccine does not work. This claim is about the effectiveness of the vaccine in preventing hospitalizations.}

\textit{2. The social media post shares information from a friend who works in a retirement home in Graz. It mentions that 75 residents were vaccinated, and 9 are still alive. This post appears to be making a claim about the vaccine's effectiveness in preventing deaths, specifically in a retirement home setting.}

\textit{Now, let's analyze the relevance:}

\textit{- Both the debunked claim and the social media post are related to the effectiveness of the Covid-19 vaccine.}

\textit{- However, the debunked claim focuses on hospitalizations, while the social media post focuses on deaths in a specific setting (a retirement home).}

\textit{- The post does not mention hospitalizations, and the debunked claim does not mention deaths in a retirement home.}

\textit{Considering these points, while both the debunked claim and the social media post are related to vaccine effectiveness, they focus on different aspects (hospitalizations vs. deaths in a specific setting). However, the core idea of questioning vaccine effectiveness is present in both.}

\textit{Answer: Yes}

\subsubsection{Example \#5 - Missing Context}

\noindent
\textbf{Post (En):} \textit{A message from a month ago at news.cafe.....}

\noindent
\textbf{Fact-check (En):} \textit{Photo of Dutch protests is current}

\noindent
\textbf{Ground truth relevance:} No

\noindent
\textbf{Model:} \texttt{C4AI Command R+} using \textbf{CoT} prompting

\noindent
\textbf{Generated response:} \textit{Reasoning: The claim and the social media post refer to different time frames. While the claim suggests that the photo is from a recent event, the post indicates that the message is from a month ago.}

\textit{Answer: Yes}

\subsubsection{Example \#6 - Contradiction}

\noindent
\textbf{Post (En):} \textit{RIP THE NURSE vaccinated yesterday DIED. Media and politicians try to hide the truth and desperately prevent the grieving family from getting the truth out to the whole world. We have clear proof that their vaccines are made to KILL DEPOPULATION. All the politicians, MEDIA, doctors, etc. - are UNFORTUNATE CRIMINALS, who campaign for these vaccines.. Make criminal complaints against organized criminal group of organized crime and genocide, qualified premeditated murder, abuse of office This nurse did kill her colleague manlawfully. Was it reckless? Is this injection really safe? Jane's Post update you've all been wa you the to share Not much and probably not what you were wanting to hear he will let us know the The questions can an Caucasian female Time of proximately to 10 hours he thanks all of you for your condolences so bad Me too. My mom doesn't want her name released to the public. They're going to do an to us within 48 hours. She wanted to be cremated so we're doing a small family ceremony. ll keep you updated but it seems my family doesn't want the publicity or her name remembered like that. Conectează-te Ninnyd Waiting~4~t... 10 h Sadly I'm hearing the nurse who collapsed who took the vaccine has since passed away, it seems social media is trying to stop the family from speaking out, yet they are trying to non the less to warn people, I hope this is wrong but from what I've seen it looks to be true. RIP HET US KNOW ENE BUTopay reauna curtir sure they'll say it wasn't vaccine related in the The questions I can answer are Age: 42, Caucasian female. Time of death approximately 8 to 10 hours after vaccine Înregistrează-te She thanks all of you for your condolences. so bad This nurse did her colle maiatulty, whitvélang Is this injection really sale Me too. My mom doesn't want her name released to the public. They're going to do an autopsy and release her body to us within 48 hours. She wanted to be cremated so we're doing a small family ceremonukeen vuundated Poți răspunde la această conversaţie Ninnyd Waiting~4~... 20 m}

\noindent
\textbf{Fact-check (En):} \textit{Spike proteins in Covid-19 mRNA vaccines are toxic.}

\noindent
\textbf{Ground truth relevance:} No

\noindent
\textbf{Model:} \texttt{Mistral 7B} with \textbf{CoT} prompting with \textbf{English} translations

\noindent
\textbf{Generated response:} 
\textit{The social media post is expressing condolences for a nurse who died after receiving a vaccine, and questioning the safety of the vaccine. The debunked claim is that spike proteins in Covid-19 mRNA vaccines are toxic.}

\textit{While the post does not explicitly mention spike proteins, it does question the safety of the vaccine, which is a common concern raised in relation to the debunked claim. Therefore, it is possible to infer that the debunked claim about spike proteins being toxic could be relevant to the post.}

\textit{However, it is important to note that the post does not provide evidence to support the claim that the vaccine caused the nurse's death. The claim that the vaccine is made to kill people is not supported by any evidence and is a conspiracy theory.}

\textit{Answer: Yes, but with a significant caveat that the claim is not supported by evidence presented in the post.}

\begin{table*}[]
\centering
\resizebox{\textwidth}{!}{%
\begin{tabular}{l|ccccc|cccc}
\toprule
\multicolumn{1}{c|}{\multirow{2}{*}{\textbf{Model}}} & \multicolumn{5}{c|}{\textbf{Original}} & \multicolumn{4}{c}{\textbf{English}} \\
\cline{2-10} 
& Zero-Shot & \makecell[c]{Zero-Shot +\\Task Description} & \makecell[c]{Few-Shot +\\Task Description} & CoT & XLT & Zero-Shot & \makecell[c]{Zero-Shot +\\Task Description} & \makecell[c]{Few-Shot +\\Task Description} & CoT \\
\midrule
All irrelevant & 45.64 & 45.64 & 45.64 & 45.64 & 45.64 & 45.64 & 45.64 & 45.64 & 45.64 \\
All relevant & 13.83 & 13.83 & 13.83 & 13.83 & 13.83 & 13.83 & 13.83 & 13.83 & 13.83 \\
\midrule
\multicolumn{10}{l}{\quad \textit{Semantic similarity baseline}} \\
\midrule
\texttt{Multilingual-E5 Large} & 68.30 & 68.30 & 68.30 & 68.30 & 68.30 & 73.46 & 73.46 & 73.46 & 73.46\\
\texttt{GTR-T5 Large} & 60.20 & 60.20 & 60.20 & 60.20 & 60.20 & 72.25 & 72.25 & 72.25 & 72.25 \\
\midrule
\multicolumn{10}{l}{\quad \textit{Textual entailment baseline}} \\
\midrule
\texttt{DeBERTa v3 Large} & 68.61 & 68.61 & 68.61 & 68.61 & 68.61 & 72.15 & 72.15 & 72.15 & 72.15 \\
\texttt{mDeBERTa v3 Base} & 64.65 & 64.65 & 64.65 & 64.65 & 64.65 & 63.13 & 63.13 & 63.13 & 63.13 \\
\midrule
\multicolumn{10}{l}{\quad \textit{LLMs with more than 70B parameters}} \\
\midrule
\texttt{Mistral Large 123B} & 70.14 & 73.33 & {\color[HTML]{009901} \textbf{81.88}} & \textbf{76.12} & {\color[HTML]{009901} \textbf{78.73}} & 72.04 & 76.45 & 80.31 & \textbf{76.12} \\
\texttt{C4AI Command R+ 104B} & {\color[HTML]{009901} \textbf{76.61}} & { \color[HTML]{009901}\textbf{82.02}} & 75.83 & 62.36 & 64.17 & {\color[HTML]{009901} \textbf{79.13}} & {\color[HTML]{009901} \textbf{79.31}} & 78.52 & 63.07 \\
\texttt{Qwen 2.5 72B Instruct} & 70.94 & 64.05 & 80.25 & 71.70 & 70.32 & 73.10 & 66.22 & {\color[HTML]{009901} \textbf{80.32}} & 73.01 \\
\texttt{Llama 3.1 70B Instruct} & 66.78 & 70.43 & 61.17 & 72.82 & 75.61 & 68.30 & 71.89 & 77.10 & 75.06 \\
\midrule
\multicolumn{10}{l}{\quad \textit{LLMs with less than 10B parameters}} \\
\midrule
\texttt{Llama 3.1 8B Instruct} & 48.41 & 70.81 &  45.05 & 66.20 & \textbf{68.68} & 57.48 & 74.37 & 61.00 & 70.92 \\
\texttt{Qwen 2.5 7B Instruct} & \textbf{65.57} & \textbf{77.99} & \textbf{72.09} & {\color[HTML]{009901} \textbf{78.63}} & 67.33 & 65.85 & \textbf{79.20} & \textbf{70.01} & {\color[HTML]{009901} \textbf{78.74}} \\
\texttt{Mistral v3 7B} & 63.58 & 63.96 & 56.02 & 70.68 & 62.95 & \textbf{66.84} & 67.79 & 66.16 & 70.63 \\
\midrule
\multicolumn{10}{l}{\quad \textit{Gemma3 Experiments}} \\
\midrule
\texttt{Gemma3 27B} & 62.66 & 59.93 & 72.56 & 55.95 & 50.31 & 59.68 & 60.27 & 70.77 & 54.26 \\
\bottomrule
\end{tabular}}
\caption{The Macro F1 performance across LLMs and prompting techniques for both original language input and English translations. The best performance is highlighted in \textbf{bold}, with the overall best performance for each prompting strategy marked in {\color[HTML]{009901} green}. Only \texttt{Llama3.1 8B} with few-shot prompting achieved lower performance than the baseline.}
\label{tab:macro-f1}
\end{table*}

\begin{table*}[]
\centering
\resizebox{\textwidth}{!}{%
\begin{tabular}{l|ccccc|cccc}
\toprule
\multicolumn{1}{c|}{\multirow{2}{*}{\textbf{Model}}} & \multicolumn{5}{c|}{\textbf{Original}} & \multicolumn{4}{c}{\textbf{English}} \\
\cline{2-10} 
& \textit{Zero-Shot} & \textit{\makecell[c]{Zero-Shot +\\Task Description}} & \textit{\makecell[c]{Few-Shot +\\Task Description}} & \textit{CoT} & \textit{XLT} & \textit{Zero-Shot} & \textit{\makecell[c]{Zero-Shot +\\Task Description}} & \textit{\makecell[c]{Few-Shot +\\Task Description}} & \textit{CoT} \\
\midrule
All irrelevant & 1.00 / 1.00 & 1.00 / 1.00 & 1.00 / 1.00 & 1.00 / 1.00 & 1.00 / 1.00 & 1.00 / 1.00 & 1.00 / 1.00 & 1.00 / 1.00 & 1.00 / 1.00 \\
All relevant & 0.00 / 0.00 & 0.00 / 0.00 & 0.00 / 0.00 & 0.00 / 0.00 & 0.00 / 0.00 & 0.00 / 0.00 & 0.00 / 0.00 & 0.00 / 0.00 & 0.00 / 0.00 \\
\midrule
\multicolumn{10}{l}{\quad \textit{Semantic similarity baseline}} \\
\midrule
\texttt{Multilingual-E5 Large} & 0.81 / 0.34 & 0.81 / 0.34 & 0.81 / 0.34 & 0.81 / 0.34 & 0.81 / 0.34 & 0.84 / 0.26 & 0.84 / 0.26 & 0.84 / 0.26 & 0.84 / 0.26 \\
\texttt{GTR-T5 Large} & 0.74 / 0.42 & 0.74 / 0.42 & 0.74 / 0.42 & 0.74 / 0.42 & 0.74 / 0.42 & 0.81 / 0.20 & 0.81 / 0.20 & 0.81 / 0.20 & 0.81 / 0.20 \\
\midrule
\multicolumn{10}{l}{\quad \textit{Textual entailment baseline}} \\
\midrule
\texttt{DeBERTa v3 Large} & 0.97 / 0.66 & 0.97 / 0.66 & 0.97 / 0.66 & 0.97 / 0.66 & 0.97 / 0.66 & 0.98 / 0.63 & 0.98 / 0.63 & 0.98 / 0.63 & 0.98 / 0.63 \\
\texttt{mDeBERTa v3 Base} & 0.98 / 0.75 & 0.98 / 0.75 & 0.98 / 0.75 & 0.98 / 0.75 & 0.98 / 0.75 & 0.99 / 0.79 & 0.99 / 0.79 & 0.99 / 0.79 & 0.99 / 0.79 \\
\midrule
\multicolumn{10}{l}{\quad \textit{LLMs with more than 70B parameters}} \\
\midrule
\texttt{Mistral Large 123B} & 0.73 / \textbf{0.04} & 0.77 / 0.04 & 0.91 / 0.19 & \textbf{0.80} / \textbf{0.05} & \textbf{0.85} / 0.10 & 0.76 / \textbf{0.06} & 0.80 / 0.05 & 0.87 / \textbf{0.13} & \textbf{0.81} / 0.07 \\
\texttt{C4AI Command R+ 104B} & \textbf{0.97} / 0.51 & \textbf{0.91} / 0.21 & 0.95 / 0.48 & 0.63 / 0.06 & 0.64 / \textbf{0.03} & \textbf{0.95} / 0.39 & \textbf{0.86} / 0.12 & 0.89 / 0.24 & 0.63 / \textbf{0.06} \\
\texttt{Qwen 2.5 72B Instruct} & 0.77 / 0.15 & 0.64 / \textbf{0.02} & 0.88 / \textbf{0.14} & 0.75 / \textbf{0.05} & 0.74 / 0.06 & 0.84 / 0.27 & 0.67 / \textbf{0.03} & 0.88 / 0.15 & 0.77 / \textbf{0.06} \\
\texttt{Llama 3.1 70B Instruct} & 0.70 / 0.09 & 0.73 / 0.05 & \textbf{0.97} / 0.79 & 0.79 / 0.13 & 0.84 / 0.18 & 0.72 / 0.10 & 0.75 / 0.06 & \textbf{0.94} / 0.42 & \textbf{0.81} / 0.12 \\
\midrule
\multicolumn{10}{l}{\quad \textit{LLMs with less than 10B parameters}} \\
\midrule
\texttt{Llama 3.1 8B Instruct} & 0.42 / \textbf{0.06} & 0.77 / \textbf{0.13} & 0.45 / \textbf{0.31} & 0.72 / \textbf{0.17} & \textbf{0.78} / 0.23 & 0.58 / \textbf{0.14} & 0.82 / \textbf{0.17} & 0.67 / \textbf{0.25} & 0.80 / \textbf{0.21} \\
\texttt{Qwen 2.5 7B Instruct} & \textbf{0.78} / 0.36 & \textbf{0.90} / 0.27 & \textbf{0.96} / 0.58 & \textbf{0.90} / 0.27 & 0.75 / \textbf{0.21} & \textbf{0.91} / 0.62 & \textbf{0.92} / 0.31 & \textbf{0.95} / 0.60 & \textbf{0.91} / 0.29 \\
\texttt{Mistral v3 7B} & 0.73 / 0.30 & 0.70 / 0.21 & 0.68 / 0.44 & 0.83 / 0.31 & 0.75 / 0.35 & 0.84 / 0.45 & 0.76 / 0.22 & 0.83 / 0.44 & 0.80 / 0.23 \\
\midrule
\multicolumn{10}{l}{\quad \textit{Gemma3 Experiments}} \\
\midrule
\texttt{Gemma3 27B} & 0.62 / 0.04 & 0.58 / 0.02 & 0.81 / 0.19 & 0.52 / 0.01 & 0.44 / 0.01 & 0.58 / 0.04 & 0.58 / 0.02 & 0.77 / 0.15 & 0.49 / 0.01 \\
\bottomrule
\end{tabular}}
\caption{The capabilities of LLMs in filtering irrelevant and relevant pairs using TNR (higher is better) and FNR (lower is better) metrics. Each cell is presented as \textit{TNR / FNR}, with the highest TNR and lowest FNR \textbf{bolded} for each prompting technique within each model category. \texttt{C4AI Command R+} and \texttt{Llama3.1 70B} achieved the highest true negative rate, while \texttt{Qwen2.5 72B} achieved the lowest false negative rate.} 
\label{tab:tnr-models}
\end{table*}

\begin{table*}[]
\resizebox{\textwidth}{!}{%
\begin{tabular}{l|ccccc|cccc}
\toprule
\multicolumn{1}{c|}{\multirow{2}{*}{\textbf{Model}}} & \multicolumn{5}{c|}{\textbf{Original}} & \multicolumn{4}{c}{\textbf{English}} \\
\cline{2-10} 
& \textit{Zero-Shot} & \textit{\makecell[c]{Zero-Shot +\\Task Description}} & \textit{\makecell[c]{Few-Shot +\\Task Description}} & \textit{CoT} & \textit{XLT} & \textit{Zero-Shot} & \textit{\makecell[c]{Zero-Shot +\\Task Description}} & \textit{\makecell[c]{Few-Shot +\\Task Description}} & \textit{CoT} \\
\midrule
All irrelevant & 0.00 / 0.00 & 0.00 / 0.00 & 0.00 / 0.00 & 0.00 / 0.00 & 0.00 / 0.00 & 0.00 / 0.00 & 0.00 / 0.00 & 0.00 / 0.00 & 0.00 / 0.00 \\
All relevant & 1.00 / 1.00 & 1.00 / 1.00 & 1.00 / 1.00 & 1.00 / 1.00 & 1.00 / 1.00 & 1.00 / 1.00 & 1.00 / 1.00 & 1.00 / 1.00 & 1.00 / 1.00 \\
\midrule
\multicolumn{10}{l}{\quad \textit{Semantic similarity baseline}} \\
\midrule
\texttt{Multilingual E5 Large} & 0.66 / 0.19 & 0.66 / 0.19 & 0.66 / 0.19 & 0.66 / 0.19 & 0.66 / 0.19 & 0.74 / 0.16 & 0.74 / 0.16 & 0.74 / 0.16 & 0.74 / 0.16 \\
\texttt{GTR-T5 Large} & 0.58 / 0.26 & 0.58 / 0.26 & 0.58 / 0.26 & 0.58 / 0.26 & 0.58 / 0.26 & 0.80 / 0.19 & 0.80 / 0.19 & 0.80 / 0.19 & 0.80 / 0.19 \\
\midrule
\multicolumn{10}{l}{\quad \textit{Textual entailment baseline}} \\
\midrule
\texttt{DeBERTa v3 Large} & 0.34 / 0.03 & 0.34 / 0.03 & 0.34 / 0.03 & 0.34 / 0.03 & 0.34 / 0.03 & 0.37 / 0.02 & 0.37 / 0.02 & 0.37 / 0.02 & 0.37 / 0.02 \\
\texttt{mDeBERTa v3 Base} & 0.25 / 0.02 & 0.25 / 0.02 & 0.25 / 0.02 & 0.25 / 0.02 & 0.25 / 0.02 & 0.21 / 0.01 & 0.21 / 0.01 & 0.21 / 0.01 & 0.21 / 0.01 \\
\midrule
\multicolumn{10}{l}{\quad \textit{LLMs with more than 70B parameters}} \\
\midrule
Mistral Large 123B & \textbf{0.96} / 0.27 & 0.96 / 0.23 & 0.81 / 0.09 & \textbf{0.95} / \textbf{0.20} & 0.90 / \textbf{0.15} & \textbf{0.94} / 0.24 & 0.95 / 0.20 & \textbf{0.87} / 0.13 & 0.93 / \textbf{0.19} \\
C4AI Command R+ 104B & 0.49 / \textbf{0.03} & 0.79 / \textbf{0.09} & 0.52 / \textbf{0.05} & \textbf{0.94} / 0.37 & \textbf{0.97} / 0.36 & 0.61 / \textbf{0.05} & 0.88 / \textbf{0.14} & 0.76 / \textbf{0.11} & \textbf{0.94} / 0.37 \\
Qwen2.5 72B Instruct & 0.85 / 0.23 & \textbf{0.98} / 0.36 & \textbf{0.86} / 0.12 & \textbf{0.95} / 0.25 & 0.94 / 0.26 & 0.73 / 0.16 & \textbf{0.97} / 0.33 & 0.85 / 0.12 & 0.94 / 0.23 \\
Llama3.1 70B Instruct & 0.91 / 0.30 & 0.95 / 0.27 & 0.21 / 0.03 & 0.87 / 0.21 & 0.82 / 0.16 & 0.90 / 0.28 & 0.94 / 0.25 & 0.58 / 0.06 & 0.88 / \textbf{0.19} \\
\midrule
\multicolumn{10}{l}{\quad \textit{LLMs with less than 10B parameters}} \\
\midrule
Llama3.1 8B Instruct & \textbf{0.94} / 0.58 & \textbf{0.87} / 0.23 & \textbf{0.69} / 0.55 & \textbf{0.83} / 0.28 & 0.77 / \textbf{0.22} & \textbf{0.86} / 0.42 & \textbf{0.83} / 0.18 & \textbf{0.75} / 0.33 & \textbf{0.79} / 0.20 \\
Qwen2.5 7B Instruct & 0.64 / \textbf{0.22} & 0.73 / \textbf{0.10} & 0.42 / \textbf{0.04} & 0.73 / \textbf{0.10} & \textbf{0.79} / 0.25 & 0.38 / \textbf{0.09} & 0.69 / \textbf{0.08} & 0.40 / \textbf{0.05} & 0.71 / \textbf{0.09} \\
Mistral v3 7B & 0.70 / 0.27 & 0.79 / 0.30 & 0.56 / 0.32 & 0.69 / 0.17 & 0.65 / 0.25 & 0.55 / 0.16 & 0.78 / 0.24 & 0.56 / 0.17 & 0.77 / 0.20 \\ 
\midrule
\multicolumn{10}{l}{\quad \textit{Gemma3 Experiments}} \\
\midrule
\texttt{Gemma3 27B} & 0.96 / 0.38 & 0.98 / 0.42 & 0.81 / 0.19 & 0.99 / 0.48 & 0.99 / 0.56 & 0.96 / 0.42 & 0.98 / 0.42 & 0.85 / 0.23 & 0.99 / 0.51 \\
\bottomrule
\end{tabular}
}
\caption{The comparison of True positive rate (TPR, higher is better) and False positive rate (FPR, lower is better) metrics for each prompting technique.  Each cell is presented as \textit{TPR / FPR}, with the highest TPR and lowest FPR \textbf{bolded} for each prompting technique within each model category.}
\label{tab:tpr-fpr}
\end{table*}

\begin{table*}[t]
\centering
\resizebox{\textwidth}{!}{%
\begin{tabular}{lc|cc|cc|cc|cc|cc}
\toprule
\multicolumn{1}{c}{\multirow{2}{*}{\textbf{Model}}} & \multicolumn{1}{c|}{\multirow{2}{*}{\textbf{Version}}} & \multicolumn{2}{c|}{\textbf{Zero-Shot}} & \multicolumn{2}{c|}{\textbf{\makecell[c]{Zero-Shot +\\Task Description}}} & \multicolumn{2}{c|}{\textbf{\makecell[c]{Few-Shot +\\Task Description}}} & \multicolumn{2}{c|}{\textbf{CoT}} & \multicolumn{2}{c}{\textbf{XLT}} \\
\cline{3-12}
& & \textit{Mono} & \textit{Cross} & \textit{Mono} & \textit{Cross} & \textit{Mono} & \textit{Cross} & \textit{Mono} & \textit{Cross} & \textit{Mono} & \textit{Cross} \\
\midrule
\multicolumn{10}{l}{\quad \textit{Semantic similarity baseline}} \\
\midrule
\texttt{Multilingual E5 Large} & Og & 0.64 / 0.13 & 0.97 / 0.72 & 0.64 / 0.13 & 0.97 / 0.72 & 0.64 / 0.13 & 0.97 / 0.72 & 0.64 / 0.13 & 0.97 / 0.72 & 0.64 / 0.13 & 0.97 / 0.72 \\
 & En & 0.87 / 0.27 & 0.82 / 0.24 & 0.87 / 0.27 & 0.82 / 0.24 & 0.87 / 0.27 & 0.82 / 0.24 & 0.87 / 0.27 & 0.82 / 0.24 & 0.87 / 0.27 & 0.82 / 0.24 \\
\texttt{GTR-T5 Large} & Og & 0.50 / 0.20 & 0.96 / 0.82 & 0.50 / 0.20 & 0.96 / 0.82 & 0.50 / 0.20 & 0.96 / 0.82 & 0.50 / 0.20 & 0.96 / 0.82 & 0.50 / 0.20 & 0.96 / 0.82 \\
& En & 0.82 / 0.20 & 0.80 / 0.20 & 0.82 / 0.20 & 0.80 / 0.20 & 0.82 / 0.20 & 0.80 / 0.20 & 0.82 / 0.20 & 0.80 / 0.20 & 0.82 / 0.20 & 0.80 / 0.20 \\
\midrule
\multicolumn{10}{l}{\quad \textit{Textual entailment baseline}} \\
\midrule
\texttt{DeBERTav3 Large} & Og & 0.96 / 0.61 & 0.97 / 0.75 & 0.96 / 0.61 & 0.97 / 0.75 & 0.96 / 0.61 & 0.97 / 0.75 & 0.96 / 0.61 & 0.97 / 0.75 & 0.96 / 0.61 & 0.97 / 0.75 \\
& En & 0.98 / 0.59 & 0.98 / 0.70 & 0.98 / 0.59 & 0.98 / 0.70 & 0.98 / 0.59 & 0.98 / 0.70 & 0.98 / 0.59 & 0.98 / 0.70 & 0.98 / 0.59 & 0.98 / 0.70 \\
\texttt{mDeBERTa v3 Base} & Og & 0.97 / 0.69 & 0.98 / 0.84 & 0.97 / 0.69 & 0.98 / 0.84 & 0.97 / 0.69 & 0.98 / 0.84 & 0.97 / 0.69 & 0.98 / 0.84 & 0.97 / 0.69 & 0.98 / 0.84 \\
 & En & 0.99 / 0.75 & 0.99 / 0.86 & 0.99 / 0.75 & 0.99 / 0.86 & 0.99 / 0.75 & 0.99 / 0.86 & 0.99 / 0.75 & 0.99 / 0.86 & 0.99 / 0.75 & 0.99 / 0.86 \\
\midrule
\multicolumn{12}{l}{\quad \textit{LLMs with more than 70B parameters}} \\
\midrule 
\multirow[t]{2}{*}{\texttt{Mistral Large 123B}} & Og & 0.71 / \textbf{0.04} & 0.74 / \textbf{0.03} & 0.73 / 0.04 & 0.79 / 0.04 & 0.89 / 0.19 & 0.91 / 0.18 & \textbf{0.81} / \textbf{0.05} & \textbf{0.80} / 0.05 & \textbf{0.86} / 0.11 & \textbf{0.84} / 0.09 \\
 & En & 0.76 / \underline{0.06} & 0.75 / \underline{0.06} & 0.81 / 0.05 & 0.80 / 0.03 & 0.87 / \underline{0.14} & 0.88 / \underline{0.10} & 0.81 / 0.07 & \underline{0.81} / 0.06 & - & - \\
\multirow[t]{2}{*}{\texttt{C4AI Command R+ 104B}} & Og & \textbf{0.96} / 0.42 & \textbf{0.98} / 0.66 & \textbf{0.89} / 0.15 & \textbf{0.94} / 0.32 & 0.95 / 0.47 & 0.95 / 0.50 & 0.61 / \textbf{0.05} & 0.64 / 0.07 & 0.62 / \textbf{0.03} & 0.67 / \textbf{0.03} \\
 & En & \underline{0.94} / 0.38 & \underline{0.95} / 0.40 & \underline{0.86} / 0.13 & \underline{0.86} / 0.12 & 0.89 / 0.23 & 0.89 / 0.26 & 0.63 / 0.06 & 0.64 / 0.05 & - & - \\
\multirow[t]{2}{*}{\texttt{Qwen 2.5 72B Instruct}} & Og & 0.72 / 0.12 & 0.82 / 0.20 & 0.62 / \textbf{0.02} & 0.65 / \textbf{0.01} & 0.87 / \textbf{0.15} & 0.88 / \textbf{0.12} & 0.75 / \underline{0.06} & 0.75 / \textbf{0.03} & 0.73 / 0.07 & 0.75 / 0.05 \\
 & En & 0.84 / 0.24 & 0.84 / 0.32 & 0.67 / \underline{0.03} & 0.67 / \underline{0.02} & 0.87 / 0.15 & 0.89 / 0.16 & 0.78 / 0.07 & 0.76 / \underline{0.04} & - & - \\
\multirow[t]{2}{*}{\texttt{Llama 3.1 70B Instruct}} & Og & 0.70 / 0.09 & 0.70 / 0.09 & 0.71 / 0.06 & 0.75 / 0.04 & \textbf{0.97} / 0.79 & \textbf{0.98} / 0.80 & 0.79 / 0.14 & 0.79 / 0.12 & 0.83 / 0.17 & \textbf{0.84} / 0.20 \\
 & En & 0.73 / 0.11 & 0.72 / 0.09 & 0.76 / 0.07 & 0.75 / 0.05 & \underline{0.94} / 0.44 & \underline{0.94} / 0.38 & \underline{0.82} / 0.12 & \underline{0.81} / 0.12 & - & - \\
\midrule
\multicolumn{12}{l}{\quad \textit{LLMs with less than 10B parameters}} \\
\midrule 
\multirow[t]{2}{*}{\texttt{Llama 3.1 8B Instruct}} & Og & 0.38 / \textbf{0.07} & 0.46 / \textbf{0.04} & 0.73 / \textbf{0.13} & 0.80 / \textbf{0.13} & 0.47 / \textbf{0.35} & 0.44 / \textbf{0.24} & 0.71 / \textbf{0.16} & 0.73 / \textbf{0.18} & \textbf{0.78} / 0.23 & 0.77 / \textbf{0.24} \\
 & En & 0.58 / \underline{0.13} & 0.59 / \underline{0.15} & 0.82 / \underline{0.17} & 0.83 / \underline{0.17} & 0.68 / \underline{0.24} & 0.67 / \underline{0.25} & 0.80 / \underline{0.20} & 0.79 / \underline{0.23} & - & - \\
\multirow[t]{2}{*}{\texttt{Qwen 2.5 7B Instruct}} & Og & \textbf{0.73} / 0.32 & \textbf{0.83} / 0.43 & \textbf{0.89} / 0.24 & \textbf{0.91} / 0.33 & \textbf{0.96} / 0.60 & \textbf{0.96} / 0.56 & \textbf{0.90} / 0.26 & \textbf{0.90} / 0.30 & 0.70 / \textbf{0.20} & \textbf{0.80} / \textbf{0.24} \\
 & En & \underline{0.92} / 0.61 & \underline{0.91} / 0.63 & \underline{0.92} / 0.30 & \underline{0.92} / 0.32 & \underline{0.94} / 0.61 & \underline{0.95} / 0.56 & \underline{0.91} / 0.28 & \underline{0.91} / 0.32 & - & - \\
\multirow[t]{2}{*}{\texttt{Mistral v3 7B}} & Og & 0.72 / 0.30 & 0.75 / 0.30 & 0.68 / 0.18 & 0.73 / 0.27 & 0.67 / 0.43 & 0.69 / 0.45 & 0.83 / 0.31 & 0.83 / 0.30 & 0.74 / 0.33 & 0.75 / 0.38 \\
 & En & 0.84 / 0.44 & 0.84 / 0.46 & 0.77 / 0.22 & 0.76 / 0.23 & 0.84 / 0.44 & 0.82 / 0.43 & 0.81 / 0.22 & 0.79 / 0.24 & - & - \\
\midrule
\multicolumn{10}{l}{\quad \textit{Gemma3 Experiments}} \\
\midrule
\multirow[t]{2}{*}{\texttt{Gemma3 27B}} & Og & 0.59 / 0.04 & 0.65 / 0.04 & 0.55 / 0.02 & 0.60 / 0.01 & 0.79 / 0.18 & 0.83 / 0.21 & 0.48 / 0.01 & 0.55 / 0.01 & 0.39 / 0.01 & 0.47 / 0.01 \\
& En & 0.58 / 0.04 & 0.58 / 0.04 & 0.58 / 0.02 & 0.58 / 0.01 & 0.77 / 0.16 & 0.77 / 0.12 & 0.47 / 0.01 & 0.51 / 0.01 & - & - \\
\bottomrule
\end{tabular}
}
\caption{The comparison of TNR (higher is better) and FNR (lower is better) metrics for each prompting technique in monolingual and cross-lingual settings. The best results (highest TNR and lowest FNR) for the original language are \textbf{bolded}, and English translations are \underline{underlined} for each category of LLMs (10B- vs. 70B+).}
\label{tab:cross-lingual-comparison-tnr}
\end{table*}

\end{document}